\documentclass[10pt,twocolumn,letterpaper]{article}

\usepackage{iccv}
\usepackage{times}
\usepackage{epsfig}
\usepackage{graphicx}
\usepackage{amsmath}
\usepackage{amssymb}
\usepackage{multirow,placeins,float}
 
\usepackage[pagebackref=true,breaklinks=true,letterpaper=true,colorlinks,bookmarks=false]{hyperref}
\usepackage{xr-hyper}

\iccvfinalcopy 

\def\iccvPaperID{2136} 

\ificcvfinal\fi

\author{Oran Gafni\quad Lior Wolf\quad Yaniv Taigman\\
Facebook AI Research\\
}

\usepackage{array}
\newcolumntype{x}[1]{>{\centering\arraybackslash\hspace{0pt}}p{#1}}

\usepackage{hyperref}
\usepackage{url,placeins}

\usepackage[utf8]{inputenc} 
\usepackage[T1]{fontenc}    
\usepackage{hyperref}       
\usepackage{url}            
\usepackage{booktabs,tabularx,multirow}       
\usepackage{amsfonts}       
\usepackage{nicefrac}       
\usepackage{microtype}      
\usepackage{color}
\usepackage{graphicx}
\usepackage{dsfont}
\usepackage{amsmath}

\usepackage{cuted}
\usepackage{capt-of}

\begin{document}

\title{Vid2Game: Controllable Characters Extracted from Real-World Videos}

\maketitle
\begin{strip}\centering
\vspace{-.75cm}
\includegraphics[width=\textwidth]{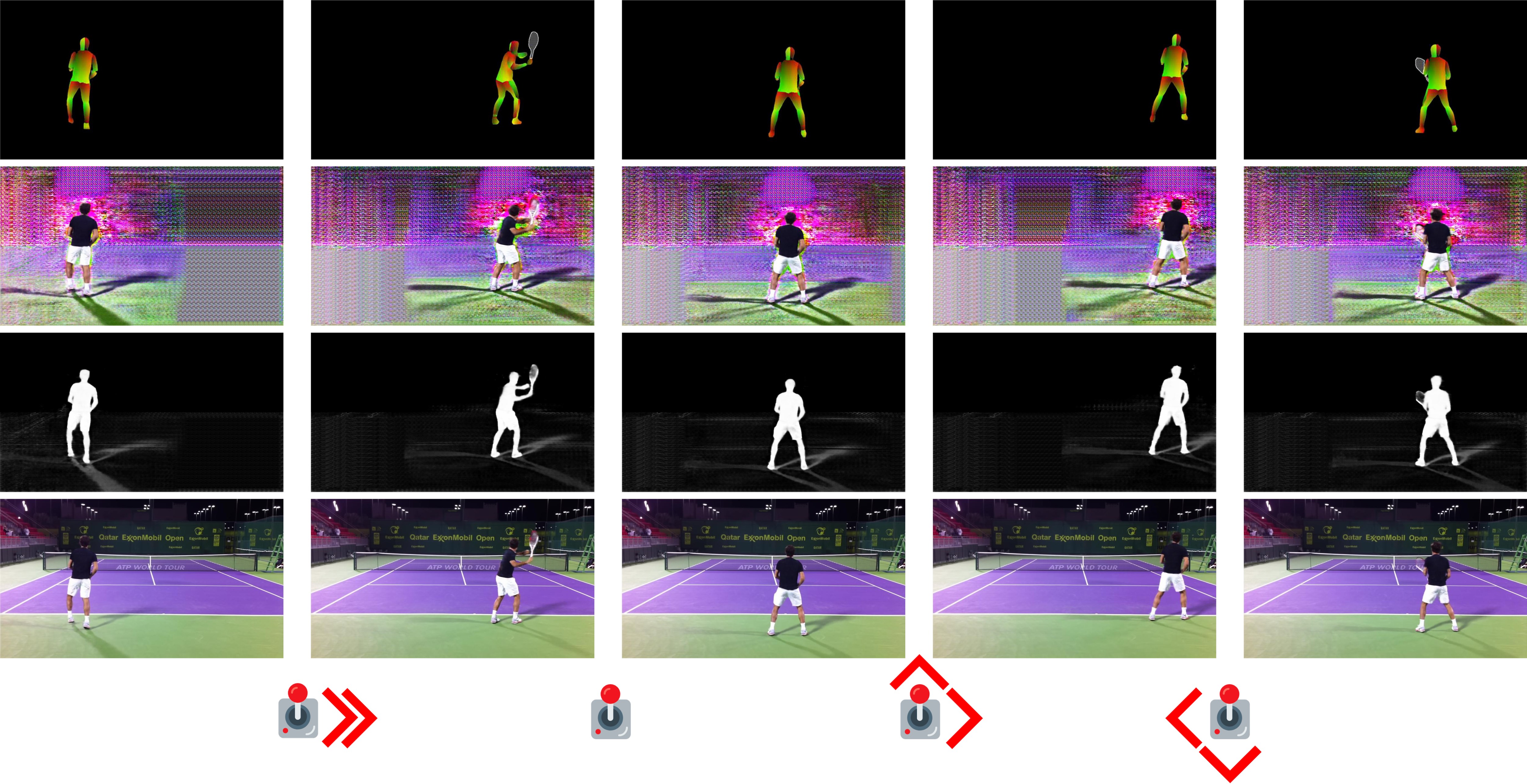}
\captionof{figure}{Our method extracts a character from an uncontrolled video and enables us to control its motion. The pose of the character, shown in the first row, is created by our Pose2Pose network in an autoregressive way, so that the motion matches the control signal illustrated by the joystick. The second row depicts the character's appearance, as generated by the Pose2Frame network, which also generates the masks shown in the third row. The final frame (last row) blends a given background and the generated frames, in accordance with these masks.}
\label{fig:teaser}
\end{strip}
\begin{abstract}
\vspace{-.5cm}
We are given a video of a person performing a certain activity, from which we extract a controllable model. The model generates novel image sequences of that person, according to arbitrary user-defined control signals, typically marking the displacement of the moving body. The generated video can have an arbitrary background, and effectively capture both the dynamics and appearance of the person. 

The method is based on two networks. The first network maps a current pose, and a single-instance control signal to the next pose. The second network maps the current pose, the new pose, and a given background, to an output frame. Both networks include multiple novelties that enable high-quality performance. This is demonstrated on multiple characters extracted from various videos of dancers and athletes.
\end{abstract}

\section{Introduction}

In this work, we propose a new video generation tool that is able to extract a character from a video, reanimate it, and generate a novel video of the modified scene, see Fig.~\ref{fig:teaser}. Unlike previous work, the reanimation is controlled by a low-dimensional signal, such as the one provided by a joystick, and the model has to complete this signal to a high-dimensional full-body signal, in order to generate realistic motion sequences. In addition, our method is general enough to position the extracted character in a new background, which is possibly also dynamic. {\color{black}A video containing a short explanation of our method, samples of output videos, and a comparison to previous work, is provided in  \url{https://youtu.be/sNp6HskavBE}}.

Our work provides a general and convenient way for human users to control the dynamic development of a given video. The input is a video, which contains one or more characters. The characters are extracted and each is associated with a sequence of displacements. In the current implementation, the motion is taken as the trajectory of the center of mass of that character in the frame. This can be readily generalized to separate different motion elements. Given a user-provided sequence (e.g., a new trajectory), a realistic video of the character, placed in front of an arbitrary background, can then be generated.

The method employs two networks, applied in a sequential manner. The first is the Pose2Pose (P2P) network, responsible for manipulating a given pose in an autoregressive manner, based on an input stream of control signals. The second is the Pose2Frame (P2F) network, accountable for generating a high-resolution realistic video frame, given an input pose and a background image.

Each network addresses a computational problem not previously fully met, together paving the way for the generation of video games with realistic graphics. The Pose2Pose network enables guided human-pose generation for a specific trained domain (e.g., a tennis player, a dancer, etc.), where guiding takes the form of 2D motion controls, while the Pose2Frame network allows the incorporation of a photo-realistic generated character into a desired environment. 

In order to enable this, we need to overcome the following challenges: (1) replacing the background requires the system to separate the character from the surroundings, which is not handled by previous work, since they either embed the character into the same learned background, or paste the generated character into the background with noticeable artifacts, (2) the separation is not binary, and some effects, such as shadow, blend the character's motion effect with that background information, (3) the control signal is arbitrary and can lead the character to poses that are not covered in the training set, and (4) generated sequences may easily drift, by accumulating small errors over time. 

Both the Pose2Pose and Pose2Frame networks adopt the pix2pixHD framework of~\cite{wang2018pix2pixHD} as the generator and discriminator backbones, yet add many contributions in order to address the aforementioned challenges.  As a building block, we use the pose representation provided by the DensePose framework~\cite{Guler2018DensePose}, unmodified.
Similarly, the hand-held object is extracted using the semantic segmentation method of ~\cite{zhou2019bottomup}, which incorporates elements from~\cite{Man+18,law2018cornernet}.

In addition to the main application of generating a realistic video from a 2D trajectory, the learned Pose2Frame network can be used for other applications. For example, instead of predicting the pose, it can be extracted from an existing video. This allows us to compare the Pose2Frame network directly with recent video-to-video solutions. 

\subsection{Problem Formulation}

The method's objective is to learn the character's motion from a video sequence, such that new videos of that character can be rendered, based on a user-provided motion sequence.

The input of the training procedure is a video sequence of a character performing an action. From this video, the pose and an approximated foreground mask are extracted by the DensePose network, augmented by the semantic segmentation of the hand-held object, for each frame. The trajectory of the center of mass is taken to be the control sequence.

At test time, the user provides a sequence of 2D displacements, and a video is created, in which the character moves in accordance with this control sequence. The background can be arbitrary, and is also selected by the user.

The method then predicts the sequence of poses based on the given control sequence (starting with an arbitrary pose), and synthesizes a video in which the character extracted from the training video is rendered in the given background.

\section{Previous Work}

\begin{figure}
  \centering
  \begin{tabular}{@{}c@{~}cc@{~}c@{~}c@{}} 
  \includegraphics[width=2cm]{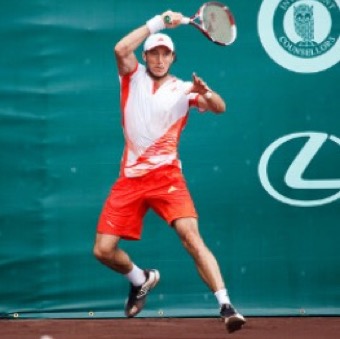} & \includegraphics[width=2cm]{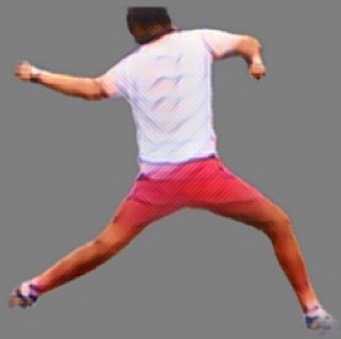} & \includegraphics[width=2cm]{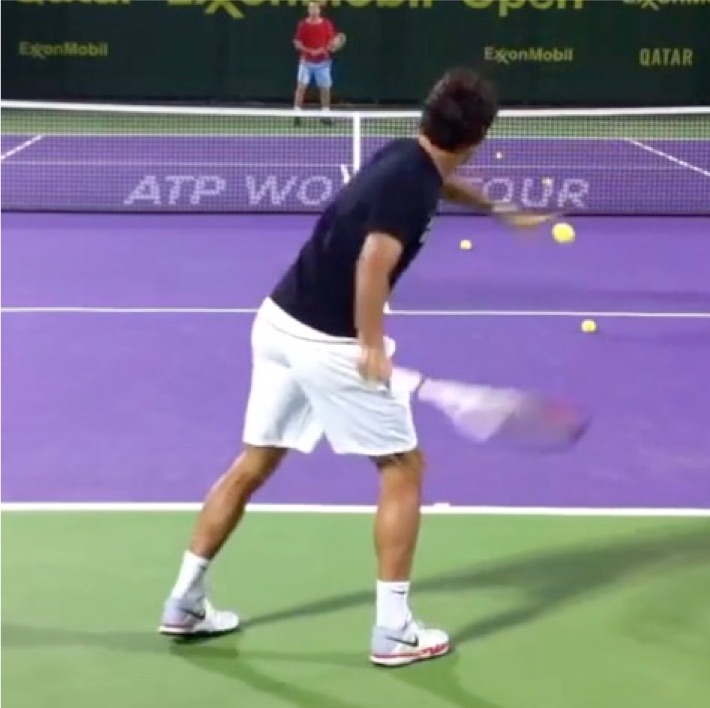} & 
  \includegraphics[width=2cm]{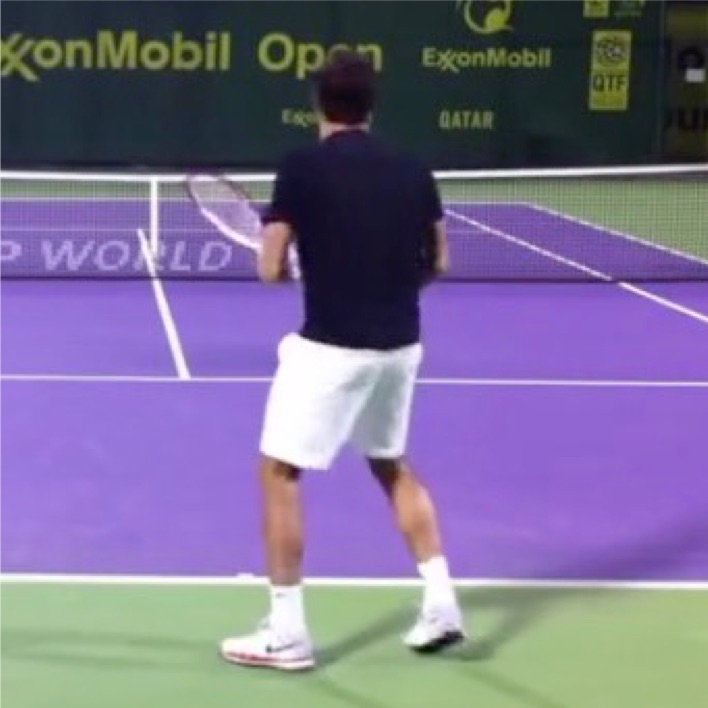} \\
  (a) & (b) & (c) & (d)
  \end{tabular}
  \caption{Comparison with~\cite{esser2018variational}. (a) The input of ~\cite{esser2018variational}, (b) the generated output of~\cite{esser2018variational}, (c) a frame from our training video, (d) our generated frame. With different objectives and dataset types, a direct comparison is not applicable. However, we present a qualitative comparison of the rendered quality for both methods.~\cite{esser2018variational} outputs a low-resolution image with noticeable artifacts, and cannot model the racket, while our result is indistinguishable from the source.} \label{fig:unet_compare}
\end{figure}

\begin{figure}
  \centering
  \begin{tabular}{@{}c@{~}cc@{~}c@{~}c@{}} 
  \includegraphics[width=2cm]{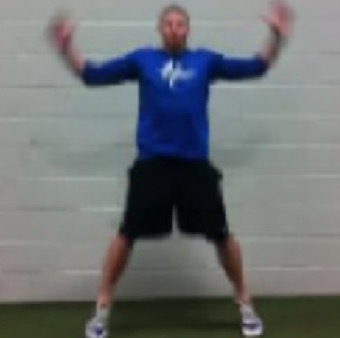} & \includegraphics[width=2cm]{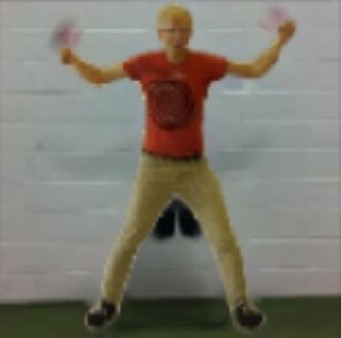} & \includegraphics[width=2cm]{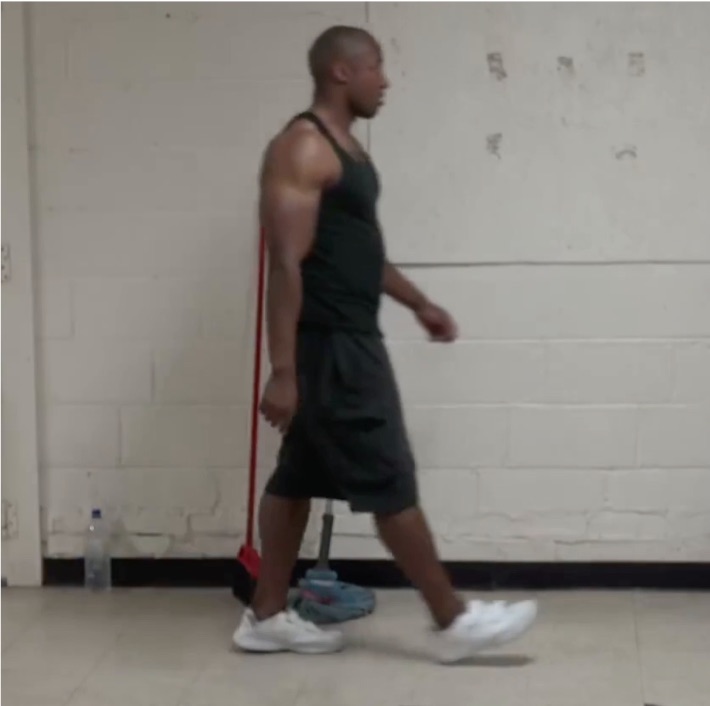} & 
  \includegraphics[width=2cm]{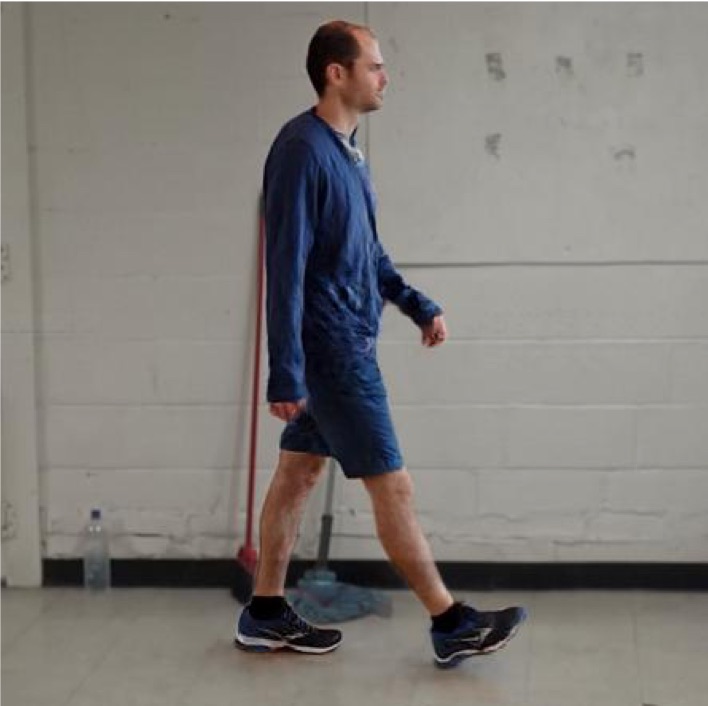} \\
  (a) & (b) & (c) & (d)
  \end{tabular}
  \caption{Comparison with~\cite{human3d}. (a) The input of ~\cite{human3d}, (b) the generated output of~\cite{human3d}, (c) our pose input, (d) P2F generated output. In contrast to our method, ~\cite{human3d} does not render environmental effects, resulting in unnatural blending of the character, undesired residues (e.g. source clothing), and works in low resolution.} \label{fig:towards_compare}
\end{figure}

There has been a vast amount of work dedicated towards video prediction and motion transfer, yet early methods were mostly fully uncontrollable or allowed domain-specific control, such as the actions of a robot arm~\cite{robotarm} or a action-conditioned Atari frame prediction~\cite{actioncondition}. 

View synthesis is a task of computer vision where unseen frames, camera views, or poses, are synthesized given a prior image. Recent approaches have also shown success in generating detailed images of human subjects in different poses ~\cite{unseen, human_dynamics18}, where some of them also condition on pose~\cite{chan2018dance, guided} to guide the generation. 

A method for learning motion patterns by analyzing YouTube videos is demonstrated in~\cite{acrobatics}, where synthetic virtual characters are set to perform complex skills in physically simulated environments, leveraging a data-driven Reinforcement Learning method that utilizes a reference motion. This method outputs a control policy that enables the character to reproduce a particular skill observed in video, which the rendered character then imitates. Unlike our method, the control signal is not provided online, one frame at a time. In addition, rendering is performed using simulated characters only, and the character in the video is not reanimated.

Autoregressive models, which can be controlled one step at a time, are suitable for the dynamic nature of video games. However, such models, including RNNs, can easily drift with long range sequences~\cite{frag}, and training RNN models for long sequences suffers from vanishing or exploding gradients. In~\cite{pfnn}, a more stable model has been proposed by generating the weights of a regression network at each frame as a function of the motion phase. However, this is mostly practical to apply given a limited number of keypoints,  whereas dense pose models contain more information.

Since the advent of Generative Adversarial Networks (GANs)~\cite{gan} and conditional GANs~\cite{mirza2014conditional}, there have been several contributions to video synthesis. In particular,~\cite{vondrick} leverage a GAN framework that separately generates the static background and the foreground motion. 

Frameworks such as vid2vid~\cite{wang2018vid2vid,chan2018dance} learn relations between different videos, and demonstrate motion transfer between faces, and from poses to body. In these contributions, the reference pose is extracted from a real frame, and the methods are not challenged with generated poses. Working with generated poses, with the accompanying artifacts and the accumulated error, is considerably more challenging. In order to address this, we incorporate a few modifications, such as relying on a second input pose, in case one of the input poses is of lesser quality, and add additional loss terms to increase the realism of the generated image. In addition, these approaches model the entire frame, including both the character and the background, which usually leads to blurry results~\cite{arbitrary, posetransfer}, particularly near the edges of the generated pose, and with complex objects, such as faces. It also leads to a loss of details from the background, and to undefinable motion of the background objects.  

A recent method that predicts video frames conditioned on spatial trajectories, has demonstrated results on moving a robotic arm~\cite{controllable}, by using dense flow maps and warping the input frame to the target. 

A method for mixing the appearance of a figure seen in an image with an arbitrary pose is presented in~\cite{esser2018variational}. While it differs greatly in the performed task, we can compare the richness of the generated images, as shown in Fig.~\ref{fig:unet_compare}. The method of~\cite{esser2018variational} results in a low-resolution output with noticeable artifacts, and cannot model the object, while our result is indistinguishable from the source. The same is true for the follow-up work~\cite{human3d}. We work at a higher video resolution of 1024p, while their work is limited to low-resolution characters, see Fig.~\ref{fig:towards_compare}.

In the same work, the authors of~\cite{human3d} also present a step toward our task  and present results for generating a controllable figure, building upon the phase-based neural network~\cite{pfnn}. This pipeline conditions on the current pose and the control signal, which we also do, using different tools: they work with extracted keypoints, while we work with the pose image. In addition, their work does not model environmental factors, such as shadows. Our work employs a mask that enables a selective modification of these environmental elements.  The videos presented by~\cite{human3d} for a controllable figure are displayed only on a synthetic background with a checkerboard floor pattern in an otherwise empty scene. These examples are limited to either walking or running, and the motion patterns are taken from an existing animation model, and present limited visual quality.

\section{Method Overview}

A video sequence with frames $f_i$ is generated, based on a sequence of poses $p_i$ and a sequence of background images $b_i$, where   $i=1,2,\dots$ is the frame index. The frame generation process also involves a sequence of spatial masks $m_i$ that determine which regions of the background are replaced by synthesized image information $z_i$.

To generate a video, the user provides the pose at time zero: $p_0$, the sequence of background images $b_i$ (which can be static, i.e., $\forall i~~ b_i=b$) and a sequence of control signals $s_i$. In our experiments, the control signal is typically comprised of the desired 2D displacement of the animated character. 

Our method is an autoregressive pose model, coupled with a frame-rendering mechanism. The first aspect of our method creates a sequence of poses, and optionally of hand-held objects. Each pose and object pair $[p_i,obj_i]$ is dependent on the previous pair $[p_{i-1},obj_{i-1}]$, as well as on the current control signal $s_i$. The second aspect generates the current frame $f_i$, based on the current background image $b_i$, the previous combined pose and object $p_{i-1}+obj_{i-1}$, and the current combined pose and object $p_i+obj_i$. The pose and object are combined by simply summing the object channel with each of the three RGB channels that encode the pose. This rendering process includes the generation of both a raw image output $z_i$ and a blending mask $m_i$. $m_i$ has values between 0 and 1, with $1-m_i$ denoting the inverted mask. 

Formally, the high-level processing is given by the following three equations:
\begin{align}
[p_i,obj_i] &= P2P([p_{i-1},obj_{i-1}],s_i)\label{eq:1}\\
(z_i,m_i) &= P2F([p_{i-1}+obj_{i-1},p_i+obj_i])\label{eq:2}\\
f_i &= z_i\odot m_i + b_i \odot (1-m_i) \label{eq:3}
\end{align}

where $P2P$ and $P2F$ are the Pose2Pose and the Pose2Frame networks. As stated, $P2F$ returns a pair of outputs that are then linearly blended with the desired background, using the per-pixel multiplication operator $\odot$.





\begin{figure*}
  \centering
 \includegraphics[width=0.9\textwidth]{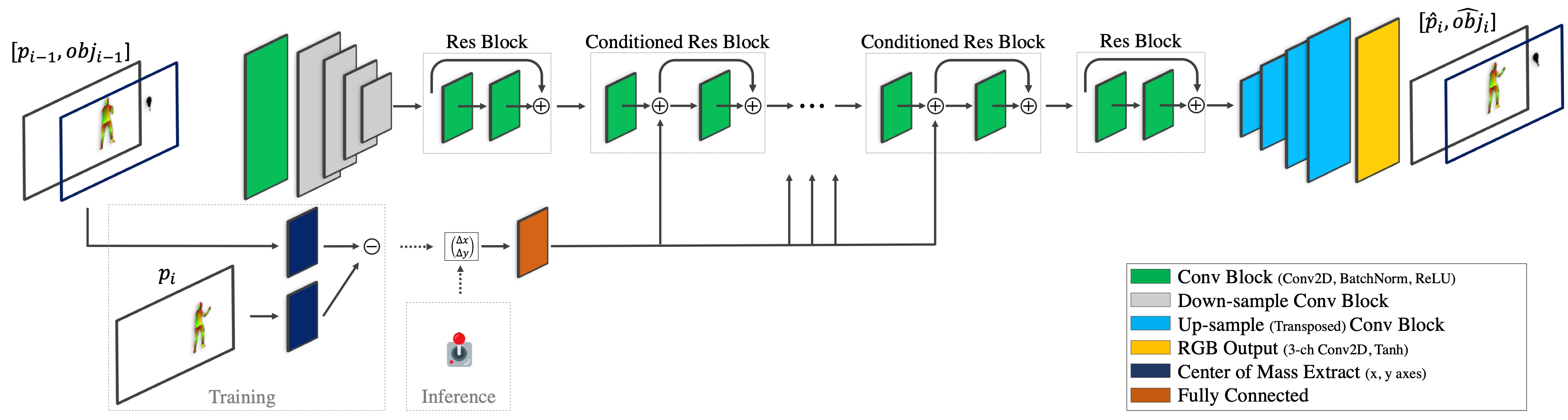} 
  \caption{The architecture of the Pose2Pose generator. During training, the middle $n_r-2$ residual blocks are conditioned by a linear projection (FC layer) of the center-mass differences between consecutive frames (in the x and y axes). For each concatenation of input pose and object [$p_{i-1},obj_{i-1}$], the network generates the next consecutive pose and object [$p_{i},obj_{i}$]. At inference time, the network generates the next pose-object pair in an autoregressive manner, conditioned on input directions.}
  \label{fig:arch_p2p}
\end{figure*}

\begin{figure*}
  \centering
    \begin{tabular}{c@{~}|@{}c} 
  \includegraphics[width=0.75\textwidth]{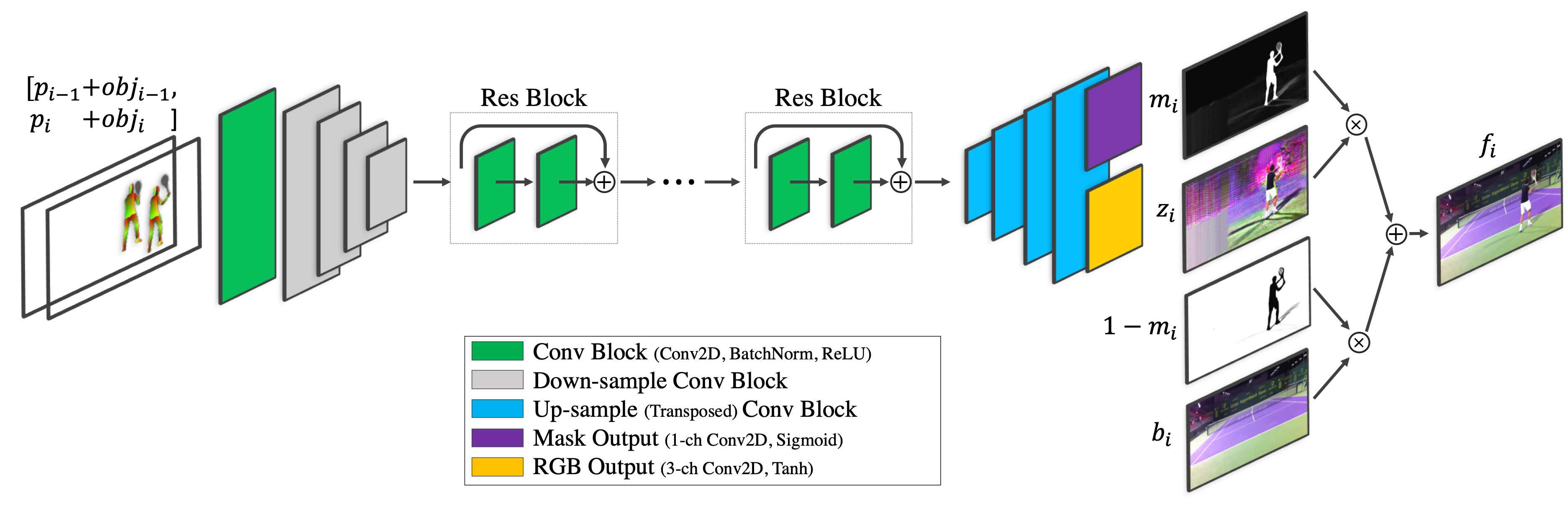} & \includegraphics[width=0.25\textwidth]{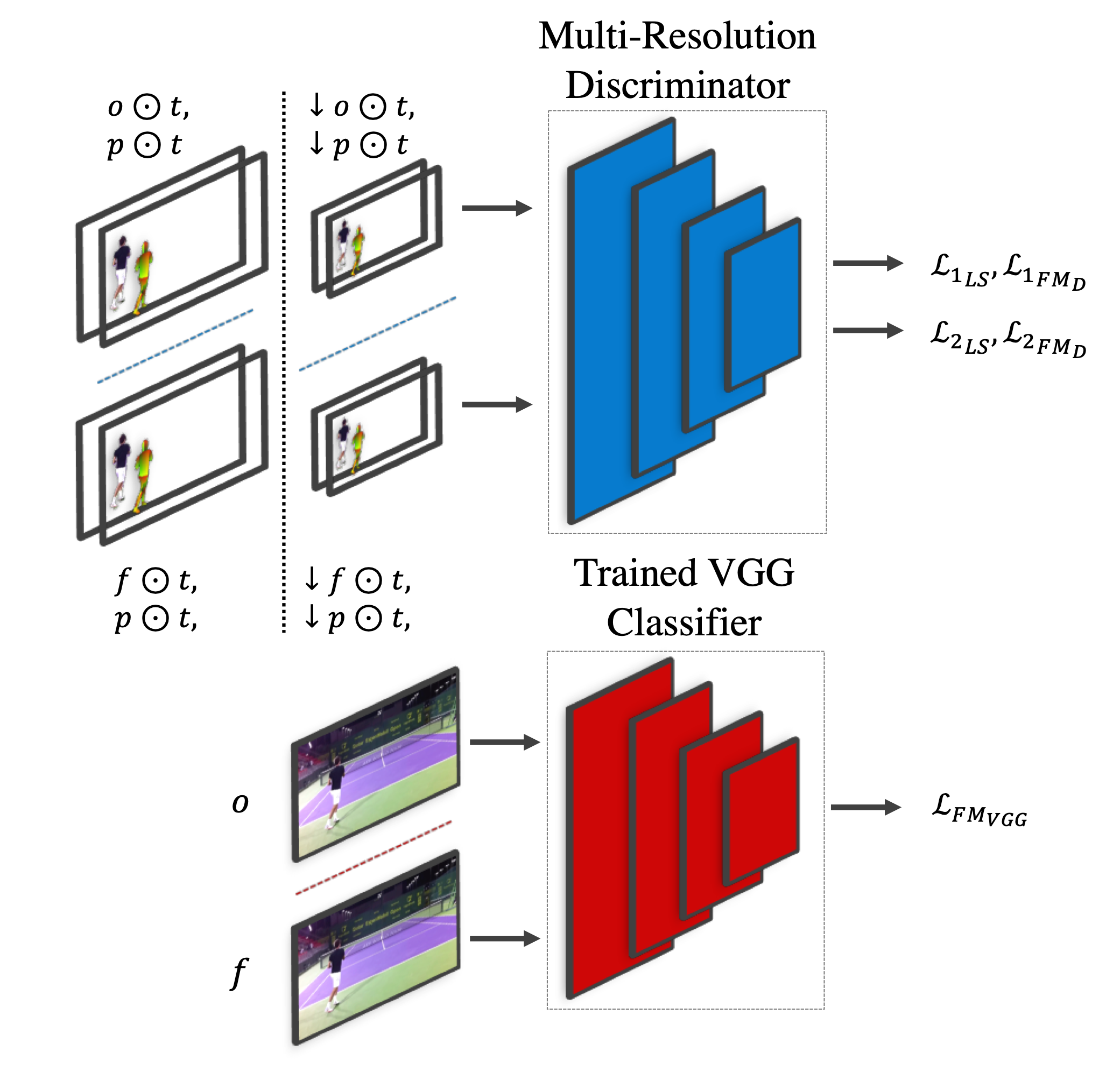} \\ 
  (a) & (b)
  \end{tabular}
\\
  \caption{The Pose2Frame network. (a) For each two combined input pose and object ($p=[p_{i-1}+obj_{i-1},p_i+obj_i]$), the network generates an RGB image ($z_i$) and a mask ($m_i$). The RGB and background images are then linearly blended by the generated mask to create the output frame $f_i$. (b) The P2F discriminator setup. The multi-scale discriminator focuses on the  binary-thresholded character, obtained with the binary mask $t$, as it appears in both the ground truth image $o$ and the output of the P2F network, for a given pose $p=(p_i,p_{i-1})$. The $\downarrow$ denotes downscaling by a factor of two, obtained by average pooling, as applied before the low-resolution discriminator.  The VGG feature-matching loss term engages with the full frame, covering perceptual context in higher abstraction layers (e.g. generated shadows).}
  
  \label{fig:arch_p2f}
\end{figure*}

\section{The Pose2Pose Network}

As mentioned, the P2P network is an evolution of the pix2pixHD architecture. Although the primary use of the pix2pixHD framework in the literature is for unconditioned image-to-image translation, we show how to modify it to enable conditioning on a control signal.

For the P2P network, we scale down the input and the generated frame size to a width of 512 pixels, allowing the network to focus on pose representation, rather than high-resolution image generation. The generation of a high-res output is deferred to the P2F network. This enables us to train the P2P network much more effectively, resulting in a stable training process that generates natural dynamics, and leads to significantly reduced inference time (post-training).

\subsection{Pose Prediction Network Architecture}

The generator's architecture is illustrated in Fig.~\ref{fig:arch_p2p}. The encoder is composed of a convolutional layer, followed by convolutions with batch normalization~\cite{ioffe2015batch} and ReLU ~\cite{nair2010rectified} activations. The latent space combines a sequence of $n_r$ residual blocks. The decoder is composed of fractional strided convolutions with instance normalization ~\cite{ulyanov2016instance} and ReLU activations, followed by a single convolution terminated by a Tanh activation for the generated frame output.

Recall that the Pose2Pose network also receives the control signal as a second input (Eq.~\ref{eq:1}). In our experiment, the control signal is a vector of dimension $n_d=2$ representing displacements along the $x$ and $y$ axes. This signal is incorporated into the network, by conditioning the center $n_r-2$ blocks of the latent space. 

The conditioning takes place by adding to the activations of each residual block, a similarly sized tensor that is obtained by linearly projecting the 2D control vector $s_i$. 

Rather than applying a conditioning block based on a traditional ResNet block, we apply a modified one that does not allow for a complete bypass of the convolutional layers. This form of conditioning increases the motion naturalness.

The specific details are as follows. The Pose2Pose network contains a down-sampling encoder $e$, a latent space transformation network $r$, and an up-sampling decoder $u$. The $r$ network is conditioned on the control signal $s$.
\begin{equation}
P2P(p,s) = u(r(e(p),s))
\end{equation}

The network $r$ contains, as mentioned, $n_r$ blocks of two types: vanilla residual blocks ($v$), and conditioned blocks $w$.
\begin{equation}
r = v \circ \underbrace{w \circ w \dots \circ w}_{n_r-2\text{~~times}} \circ v
\end{equation}

The first and last block are residual blocks of the form:
\begin{equation}
v(x) = f_2(f_1(x)) + x 
\end{equation}
where $x$ denotes the activations of the previous layer, $f_1(x)$ and $f_2(x)$ are two consecutive convolutional layers. 

The conditioned block we introduce is of the form
\begin{equation}
w(x,s) = f_2(f_1(x) + g(s)) + f_1(x) + g(s)
\end{equation}
where $s$ is a 2D displacement vector, and $g$ is a fully connected network with a number of output neurons that equals the product of the dimensions of the tensor $x$. Additional architecture and implementation details of the P2P network can be found in the appendix.

\subsection{Training the Pose Prediction Network}

Following pix2pixHD~\cite{wang2018pix2pixHD}, we employ two discriminators (low-res and high-res), indexed by $k=1,2$. During training, the LSGAN~\cite{lsgan} loss is applied to the generator and discriminator. An L1 feature-matching loss is applied over the discriminators' activations, and a trained VGG~\cite{vggvergydeep} network. The loss applied to the generator can then be formulated as:
\begin{equation}
\mathcal{L}_{P2P}=\sum_{k=1}^{2}{\left( \mathcal{L}_{LS^k} + \lambda_{D}\mathcal{L}_{FM_D^k}\right)} + \lambda_{VGG}\mathcal{L}_{FM_{VGG}}
\end{equation}
where the networks are trained with $\lambda_{D}=\lambda_{VGG}=10$.
The LSGAN generator loss is (the $obj_i$ elements are omitted for brevity): 
\begin{equation}
\mathcal{L}_{LS^k}=\mathbb{E}_{(p_{i-1},s_{i})}\left[ \left(D_k(p_{i-1},P2P(p_{i-1},s_i))-\mathds{1}\right)^2 \right] \end{equation}

The expectation is computed per mini-batch, over the input pose $p_{i-1}$ and the associated $s_i$. The discriminator's feature-matching loss compares the predicted pose with that of the generated pose, using the activations of the discriminator, and is calculated as:
\begin{equation}
\begin{split}
\mathcal{L}_{FM_D^k}=\mathbb{E}_{(p_{i-1},p_{i})}\sum_{j=1}^{M}\frac{1}{N_j}||D_k^{(j)}(p_{i-1},p_{i})- \\ D_k^{(j)}(p_{i-1},P2P(p_{i-1},s_i))||_1
\end{split}
\end{equation}

with $M$ being the number of layers, $N_j$ the number of elements in each layer, $p_{i-1}$ the input (previous) pose, $p_{i}$ the current (real) pose, $P2P(p_{i-1},s)$ the estimated pose, and $D_k^{(j)}$ the activations of discriminator $k$ in layer $j$.

The VGG feature-matching loss is calculated similarly, acting as a perceptual loss over a trained VGG classifier:
\begin{equation}
\mathcal{L}_{FM_{VGG}}=\sum_{j=1}^{M}\frac{1}{N'_j}||VGG^{(j)}(p_{i})-VGG^{(j)}(P2P(p_{i-1},s_i))||_1
\end{equation}
with $N'_j$ being the number of elements in the $j$-th layer, and $VGG^{(j)}$ the VGG classifier activations at the $j$-th layer.

The loss applied to the discriminator is formulated as:
\begin{equation}
\begin{split}
\mathcal{L}_{D^k}= & \frac{1}{2}\mathbb{E}_{(p_{i-1},s_{i})}\left[ \left(D_k(p_{i-1},P2P(p_{i-1},s_i))\right)^2 \right] + \\ 
 & \frac{1}{2}\mathbb{E}_{(p_{i-1},p_{i})}\left[ \left(D_k(p_{i-1},p_i)-\mathds{1}\right)^2 \right] 
 \raisetag{12pt}
\end{split}
\end{equation}

The training sequences are first processed by employing the DensePose network, in order to extract the pose information from each frame. This pose information takes the form of an RGB image, where the 2D RGB intensity levels are a projection of the 3D UV mapping.

By applying a binary threshold over the DensePose RGB image, we are able to create a binary mask for the character in the video. From the binary mask $t_i$ of each frame $i$, we compute the center of mass of the character $\rho_i$. The control signal during training is denoted as $s_i=\rho_i-\rho_{i-1}$. 

Due to the temporal smoothness in the videos, the difference between consecutive frames in the full frame-rate videos (30fps) is too small to observe significant motion. This results in learned networks that are biased towards motionless poses. Hence, we train with $\Delta=2$ inter-frame intervals (where $\Delta=1$ describes using consecutive frames). During inference, we sample at 30fps and apply a directional conditioning signal that has half of the average motion magnitude during training.

\noindent\textbf{Stopping criteria} 
We use the Adam optimizer~\cite{adam} with a learning rate of $2\cdot10^{-4}$, $\beta_1=0.5$ and $\beta_2=0.999$. We observe that training the P2P network does not provide for monotonic improvement in output quality. We stipulate the P2P network final model to be that which yields the lowest loss, in terms of discriminator feature-matching. While there are several losses applied while training the P2P network, the discriminator feature-matching loss is the only one that holds both motion context (i.e. receives both the previous and current pose), and information of different abstraction levels (i.e. feature-matching is applied over different levels of activations). This results in improved motion naturalness, and reduced perceptual distance.

\noindent\textbf{Random occlusions} To cope with pose detection imperfections that occasionally occur, which in turn impair the quality of the generated character, we employ a dedicated data augmentation method, in order to boost the robustness of the Pose2Pose network. A black ellipse of random size and location is added to each input pose frame within the detection bounding box, resulting in an impaired pose (see appendix Fig.~\ref{fig:training_techniques}), with characteristics that are similar to "naturally" occurring imperfections.

\section{The Pose2Frame Network}

While the original pix2pixHD network transforms an entire image to an output image of the same size from a specified domain, our Pose2Frame network transforms a pose to a character that is localized in a specific part of the output image and embedded in a given, possibly dynamic, background. This is done by both refocusing the discriminators' receptive field, and applying a learned blending mask over the raw image output. The DensePose network plays a crucial role, as it provides both the relevant image region and a prior over the blending mask.

We observe that focusing the discriminator on the character eliminates the need for feature-enhancing techniques, such as the introduction of a face-GAN, as done by~\cite{chan2018dance}), or adding a temporal loss (which is useful for reducing background motion) as done by~\cite{wang2018vid2vid}.

\subsection{Frame Generation Network Architecture}

The generator architecture is illustrated in Fig.~\ref{fig:arch_p2f}.
The P2F low-level network architecture details are somewhat similar to those of the P2P network, with the following modifications: (1) the P2F network generates frames with a resolution width of 1024, (2) no conditioning is applied, i.e., the $w$ layers are replaced by $v$ layers, (3) the network generates two outputs: the raw image data $z$ and a separate blending mask $m$, (4) the discriminators are altered to reflect the added focus, and (5) new regularization terms are added to ensure that the masking takes place at the relevant regions (Eq.~\ref{eq:mask}). 

\begin{figure}
  \centering
  \begin{tabular}{cc} 
  \includegraphics[height=2.95cm]{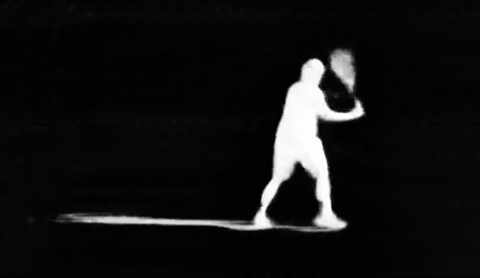} & \includegraphics[height=2.95cm]{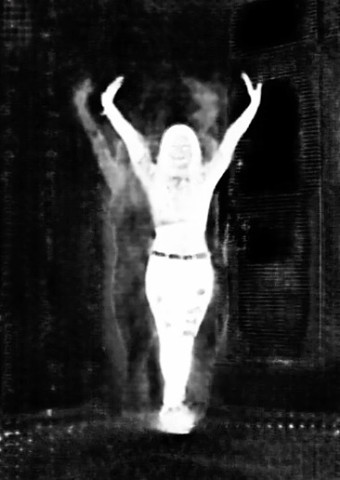} \\ 
  (a) & (b)
  \end{tabular}
  \caption{Samples of masks that model both the character and places in the scene, in which appearance is changed by the character. (a) The shadow and the tennis racket of the character are captured by the mask, (b) the dancer's shadow appears as part of the mask.}
  \label{fig:shadows}
\end{figure}

The generated mask $m$ blends the raw output $z$ with the desired background $b$, rendering the final output frame $f$, according to Eq.~\ref{eq:3} (omitting the index $i$ for brevity). Note that the blending mask is not binary, since various effects such as shadows, contain both character-derived information and background information, see Fig.~\ref{fig:shadows}. Nevertheless, we softly encourage the blending mask to favor the background in regions external to the character, and discourage the generator from rendering meaningful representations outside the character. This is done by employing several regularization terms over the generated mask. As a side effect of these added losses, the network is required to perform higher-level reasoning and not rely on memorization. In other words, instead of expanding the mask to include all background changes, the network separates between character dependent changes, such as shadows, held items, and reflections, and those that are independent. 

The discriminator setup is illustrated in Fig.~\ref{fig:arch_p2f}(b). The discriminator's attention is predominantly shifted towards the character, by applying an inverse binary mask over the character. The masked character image is fed into the discriminators, affecting both the multi-scale loss, and the feature-matching loss applied over the discriminators' activations. In parallel, the fully generated frame is fed into the VGG network, allowing the VGG feature-matching loss to aid in the generation of desired structures external to the character.

\subsection{Training the Pose to Frame Network}

The P2F generator loss is formulated as:
\begin{equation}
\mathcal{L}_{P2F}=\sum_{k=1}^{2}{\left( \mathcal{L}_{LS^k} + \lambda_{D}\mathcal{L}_{FM_D^k}\right)} + \lambda_{1}\mathcal{L}_{FM_{VGG}} + \lambda_{2}\mathcal{L}_{mask}
\end{equation}
where $\lambda_{1}=10$ and $\lambda_{2}=1$. The LSGAN generator loss is calculated as: 
\begin{equation}
\mathcal{L}_{LS^k}=\mathbb{E}_{(p,t)}\left[ \left(D_k(p\odot t,f\odot t)-1\right)^2 \right] \end{equation}
where $p=[p_{i-1}+obj_{i-1},p_i+obj_i]$ denotes the two-pose image, and $t$ is the binary mask obtained by thresholding the obtained DensePose image at time $i$.
The discriminator's feature-matching loss is calculated as:
\begin{equation}
\mathcal{L}_{FM_D^k}=\mathbb{E}_{(p,o,t)}\sum_{j=1}^{M}\frac{1}{N_j}||D_k^{(j)}(p\odot t,o\odot t)-D_k^{(j)}(p\odot t,f\odot t)||_1
\end{equation}
With $M$ being the number of layers, $N_j$ the number of elements in each layer, and $o$ the real (ground truth) frame. 
The VGG feature-matching loss is calculated over the full ground truth frame, rather than the one masked by $t$:

\begin{equation}
\mathcal{L}_{FM_{VGG}}=\sum_{j=1}^{M}\frac{1}{N_j}||VGG^{(j)}(o)-VGG^{(j)}(f)||_1
\end{equation}
with $o$ being the ground truth frame, $N_j$ being the number of elements in the $j$-th layer, and, as before, $VGG^{(j)}$ the VGG activations of the $j$-th layer.

The mask term penalizes the mask (see appendix Fig.~\ref{fig:regs} for a visual illustration):
\begin{equation}
\begin{split}
\label{eq:mask}
\mathcal{L}_{mask}=||m\odot (1-t)||_1 + ||m_x\odot (1-t)||_1 +\\ ||m_y\odot (1-t)||_1 + ||1-m\odot t||_1
\end{split}
\end{equation}
where $m$ is the generated mask, and $m_x$ and $m_y$ the mask derivatives in the x and y axes respectively. The first term acts to reduce the mask's activity outside the regions detected by DensePose. The mask, however, is required to be on in such regions, e.g., to draw shadows. Similarly, we reduce the change in the mask outside the pose-detected region, in order to eliminate secluded points there. Finally, a term is added to encourage the mask to be on in the image regions occupied by the character.

The loss applied to the two discriminators is given by:
\begin{equation}
\begin{split}
\mathcal{L}_{D^k}= & \frac{1}{2}\mathbb{E}_{(p,t)}\left[ \left(D_k(p\odot t,f\odot t)\right)^2 \right] + \\ 
 & \frac{1}{2}\mathbb{E}_{(p,o,t)}\left[ \left(D_k(p\odot t,o\odot t)-1\right)^2 \right] 
\end{split}
\end{equation}

The Adam optimizer is used for P2F similar to the P2P. The training progression across the epochs is visualized in the appendix (Fig.~\ref{fig:training_seq}).

\begin{table}[t]
\begin{center}
\begin{small}
\centering
  \begin{tabular}{l@{~~~}l@{~}c@{~}c@{~}c@{~}c}
    \toprule
    Dataset & Method & SSIM & LPIPS      & LPIPS  & LPIPS \\
     & & & (SqzNet)      & (AlexNet) & (VGG)\\
    \midrule
    \multirow{ 2}{*}{Tennis} & ours & \textbf{240}$\pm{2}$ & \textbf{265}$\pm{3}$ & \textbf{400}$\pm{4}$ & \textbf{474}$\pm{5}$ \\
     &pix2pixHD &  301$\pm{26}$ & 379$\pm{35}$ & 533$\pm{42}$ & 589$\pm{32}$ \\
    \midrule
    \multirow{ 2}{*}{Walking} & ours  & \textbf{193}$\pm{133}$  & \textbf{216}$\pm{149}$ & \textbf{365}$\pm{252}$ & \textbf{374}$\pm{258}$ \\
     & pix2pixHD  & 224$\pm{156}$  & 308$\pm{224}$ & 485$\pm{347}$ & 434$\pm{303}$  \\
    \midrule
    \multirow{2}{*}{Fencing} &Ours  & \textbf{45}$\pm{4}$  & \textbf{41}$\pm{8}$ & \textbf{52}$\pm{11}$ & \textbf{150}$\pm{15}$  \\
     & pix2pixHD  & 308$\pm{95}$  & 531$\pm{129}$ & 670$\pm{168}$ & 642$\pm{86}$  \\
  \midrule
  \end{tabular}
    \end{small}
    \end{center}
  \caption{Comparison of the P2F network with the pix2pixHD method of~\cite{wang2018pix2pixHD} {\color{black}(see also Fig.~\ref{fig:pix2pixhd})}. The distance in SSIM and LPIPS  
  from the ground-truth test set is shown for three scenarios: (1) tennis (ground truth contains dynamic elements, such as other players, crowd, slight difference in camera angle), (2) walking (different character clothing, background lighting, and camera angle), (3) fencing (same character, background, and camera angle). The results are multiplied by a factor of 1000 for readability.}
  \label{tab:pix2pixhd_comp}
\end{table}

\begin{table}[t]
\begin{small}
\centering
  \begin{tabular}{l@{~}c@{~}c@{~}c@{~}c}
    \toprule
    Network Component & SSIM & LPIPS   & LPIPS  & LPIPS \\
     & &(SqzNet)  & (AlexNet) & (VGG)\\
    \midrule
    Base Conditioning & 15.0$\pm${4} & 20.5$\pm{14}$ & 39.8$\pm{25}$ & 37.0$\pm${14} \\
    + Conditioning Block &  14.7$\pm${3} & 15.6$\pm${7} & 29.8$\pm${14} & 30.6$\pm${8} \\
    + Stopping Criteria  & 14.0$\pm${3}  & 14.9$\pm${7} & 28.1$\pm${14} & 29.5$\pm${8} \\
    + Object Channel  & 14.1$\pm${3}  & 13.3$\pm${6} & 24.9$\pm${12} & 28.6$\pm${7}  \\
    \midrule
  \end{tabular}
    \end{small}
  \caption{Ablation study of the P2P network on the tennis sequence. The results are multiplied by a factor of 1000 for readability.}
  \label{tab:ablation}
\end{table}

\begin{figure}
  \includegraphics[width=0.995\linewidth]{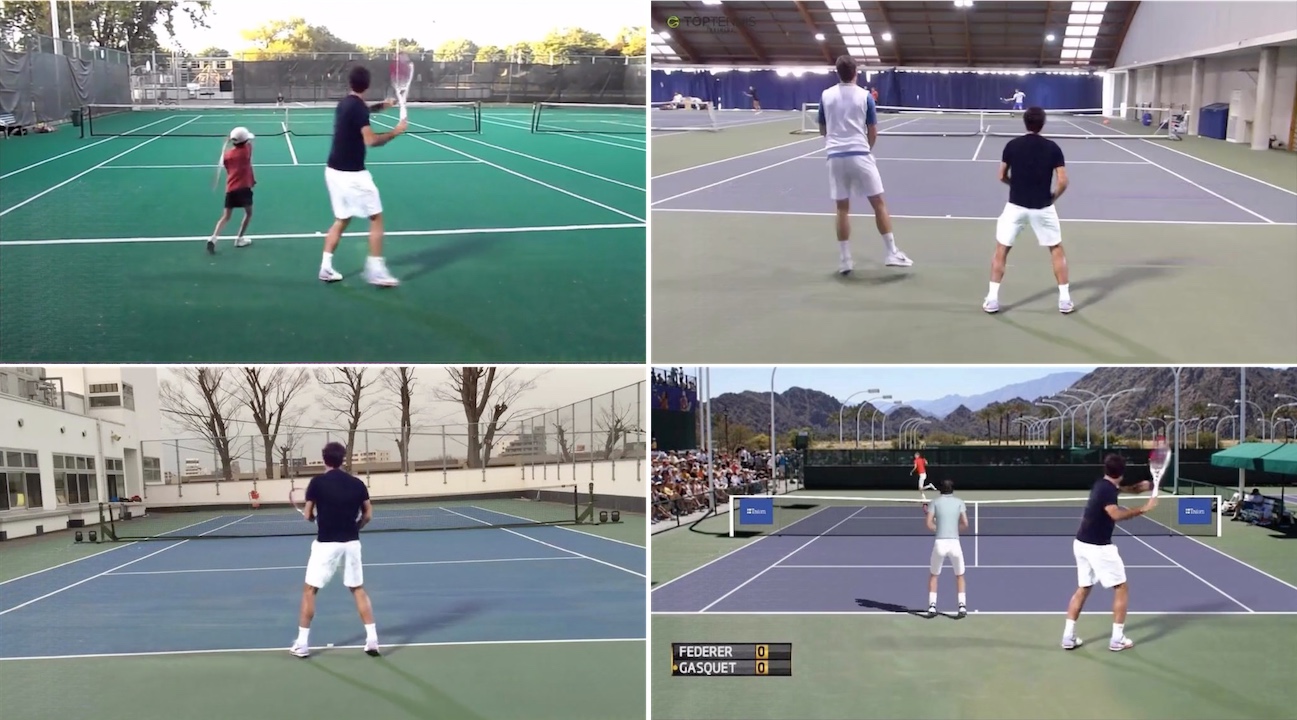}
  \caption{{\color{black}Generated frames for the controllable tennis character blended into different dynamic backgrounds.}}
  \label{fig:backgrounds_tennis}
\end{figure}

\begin{figure}
\centering
  \includegraphics[width=0.495\textwidth]{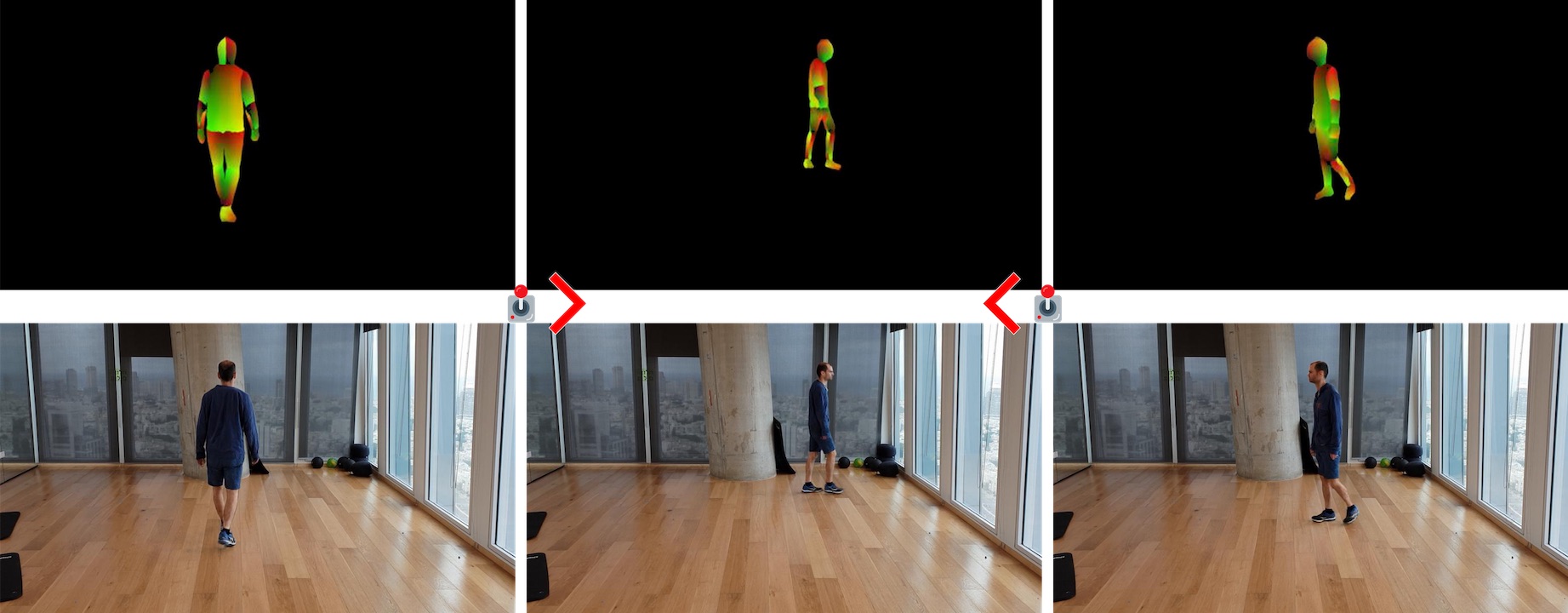}
  \caption{Synthesizing a walking character, emphasizing the control between the frames. Shown are the sequence of poses generated in an autoregressive manner, as well as the generated frames.}
  \label{fig:walk_v2}
\end{figure}

\begin{figure*}
    \centering
\resizebox{.95\textwidth}{!}{
    \begin{tabular}{ccccc}
    \includegraphics[width=.1821345234\linewidth]{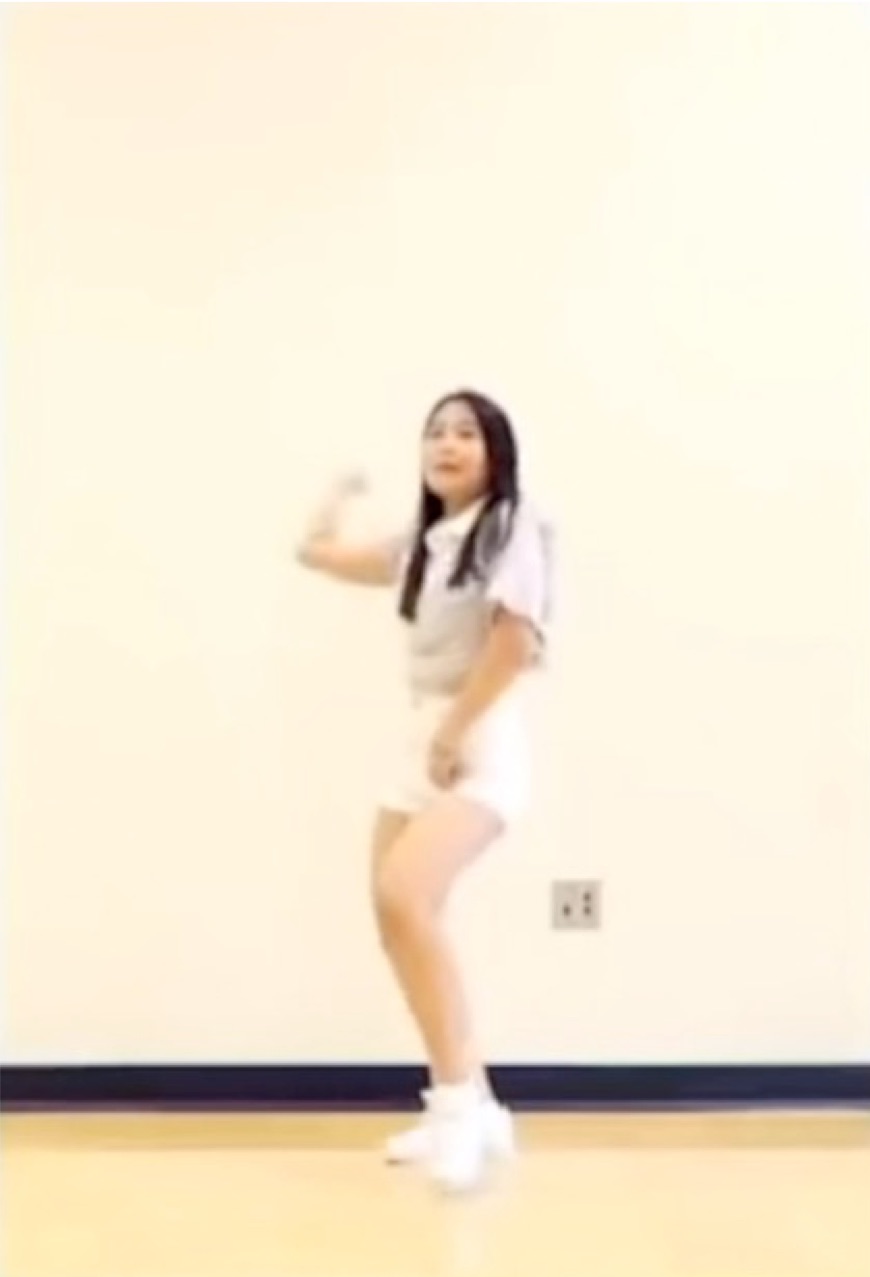}&
    \includegraphics[width=.1821345234\linewidth]{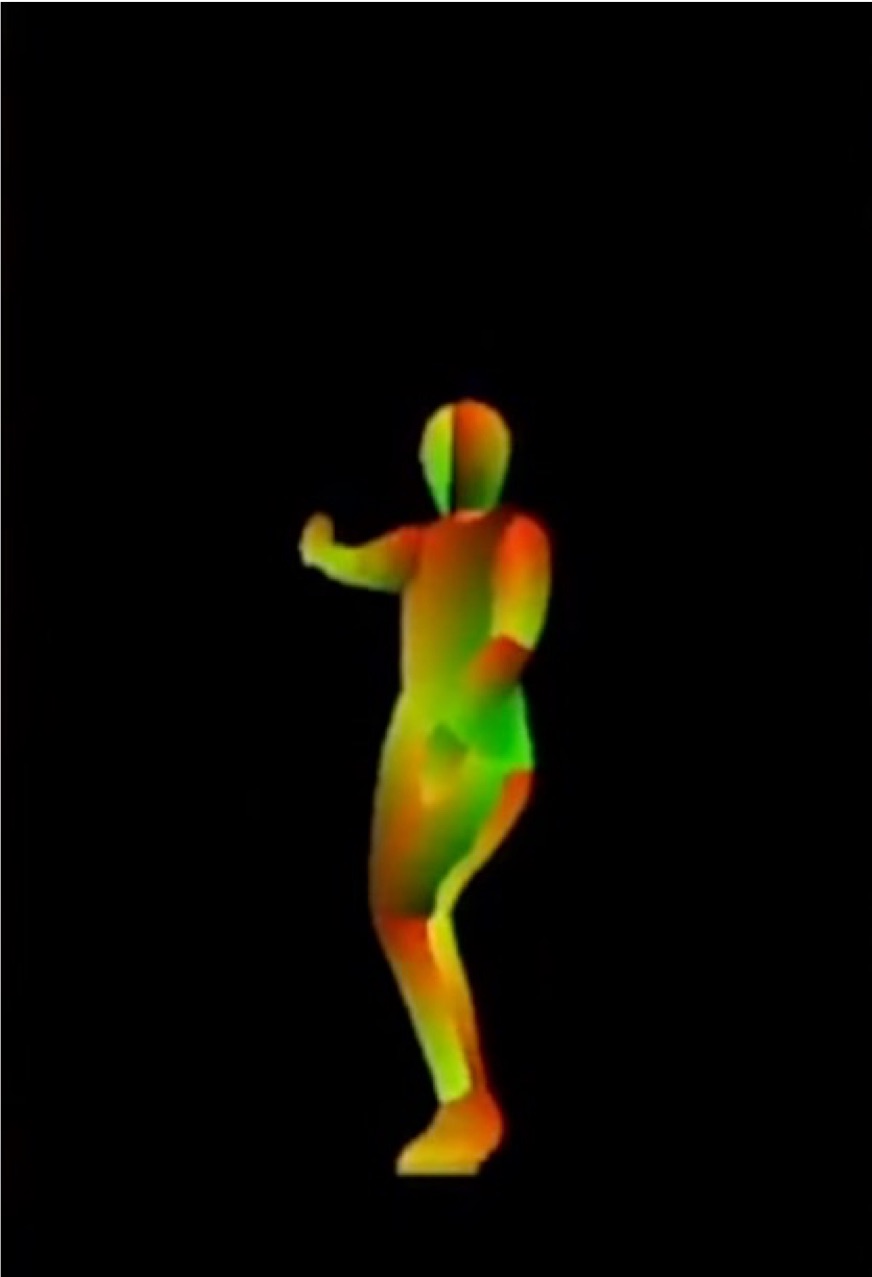}& 
    \includegraphics[width=.1821345234\linewidth]{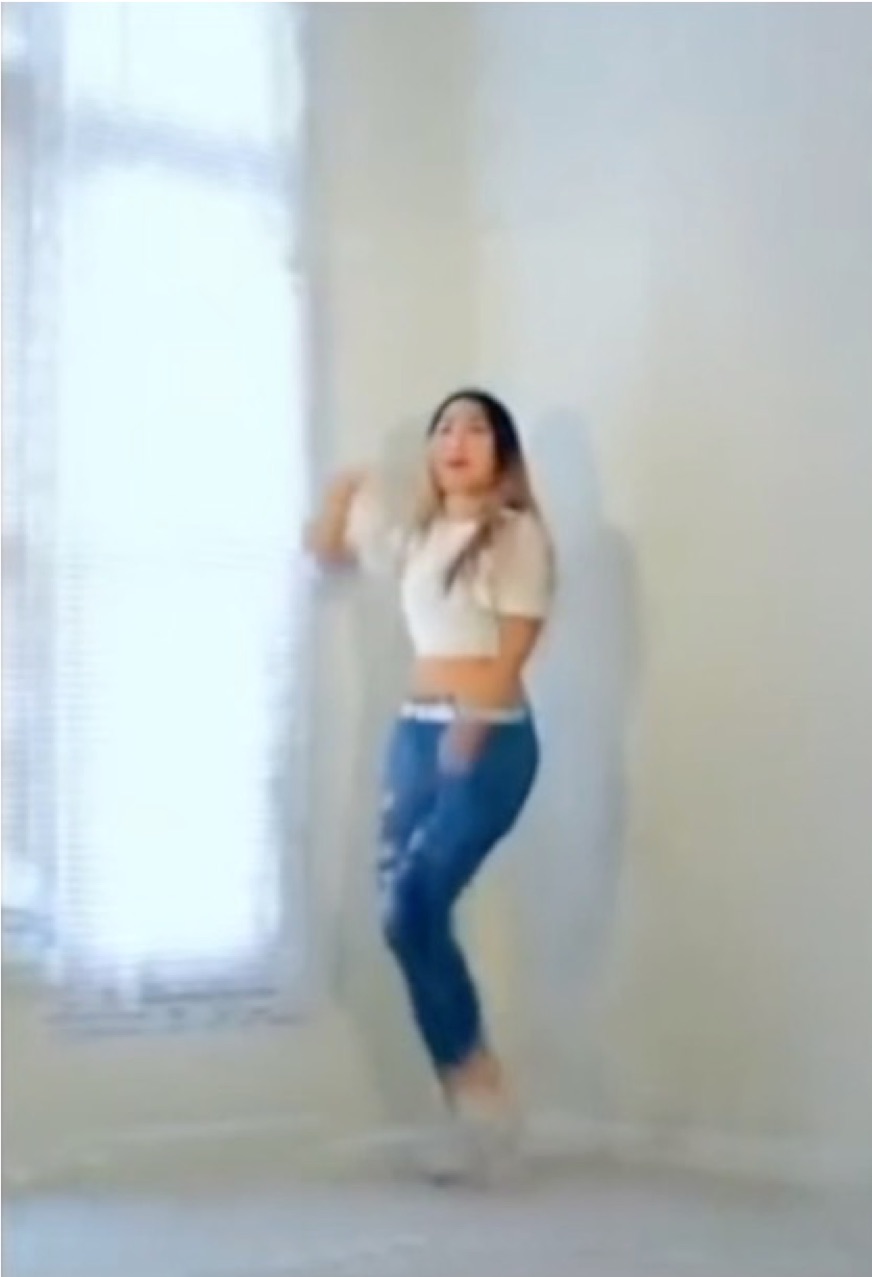}& 
    \includegraphics[width=.1821345234\linewidth]{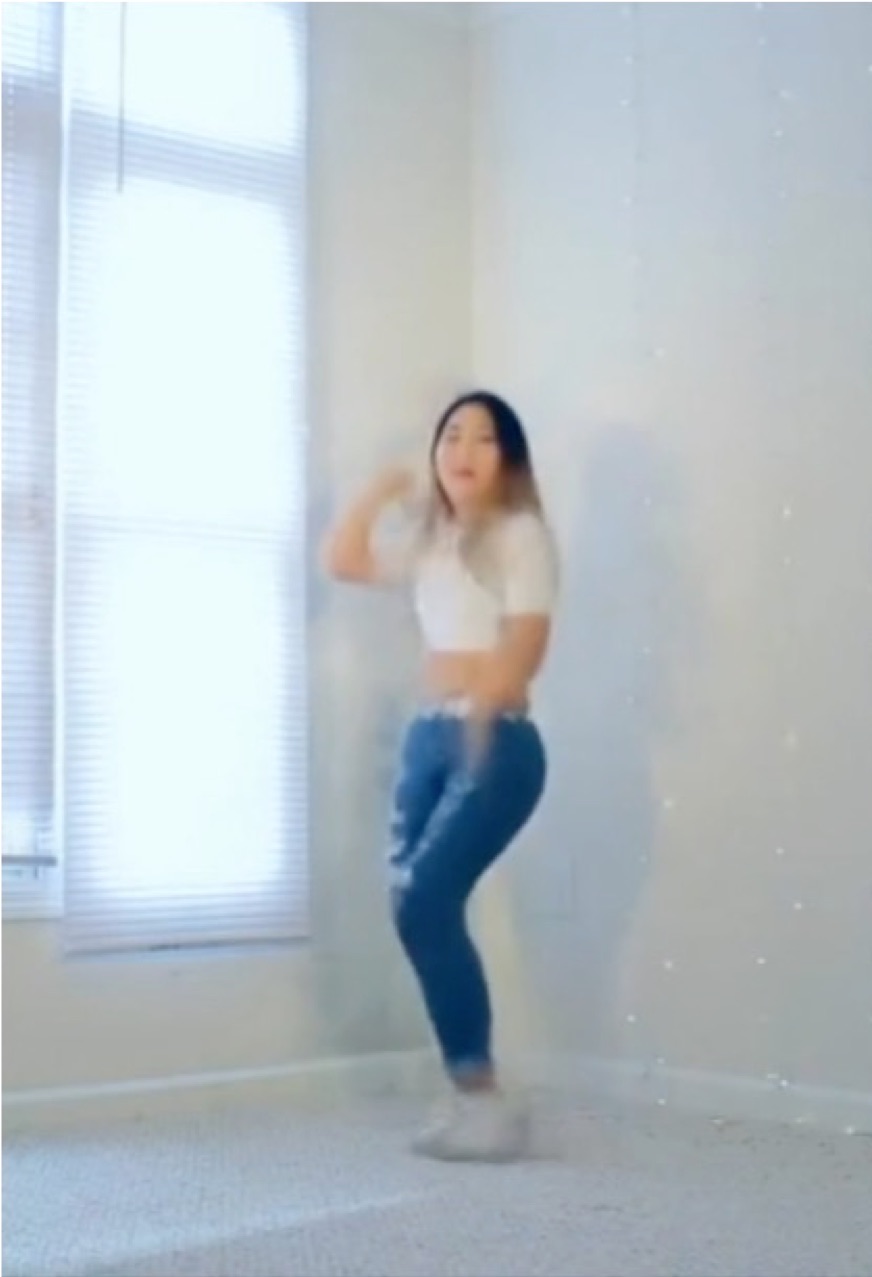}&
    \includegraphics[width=.1821345234\linewidth]{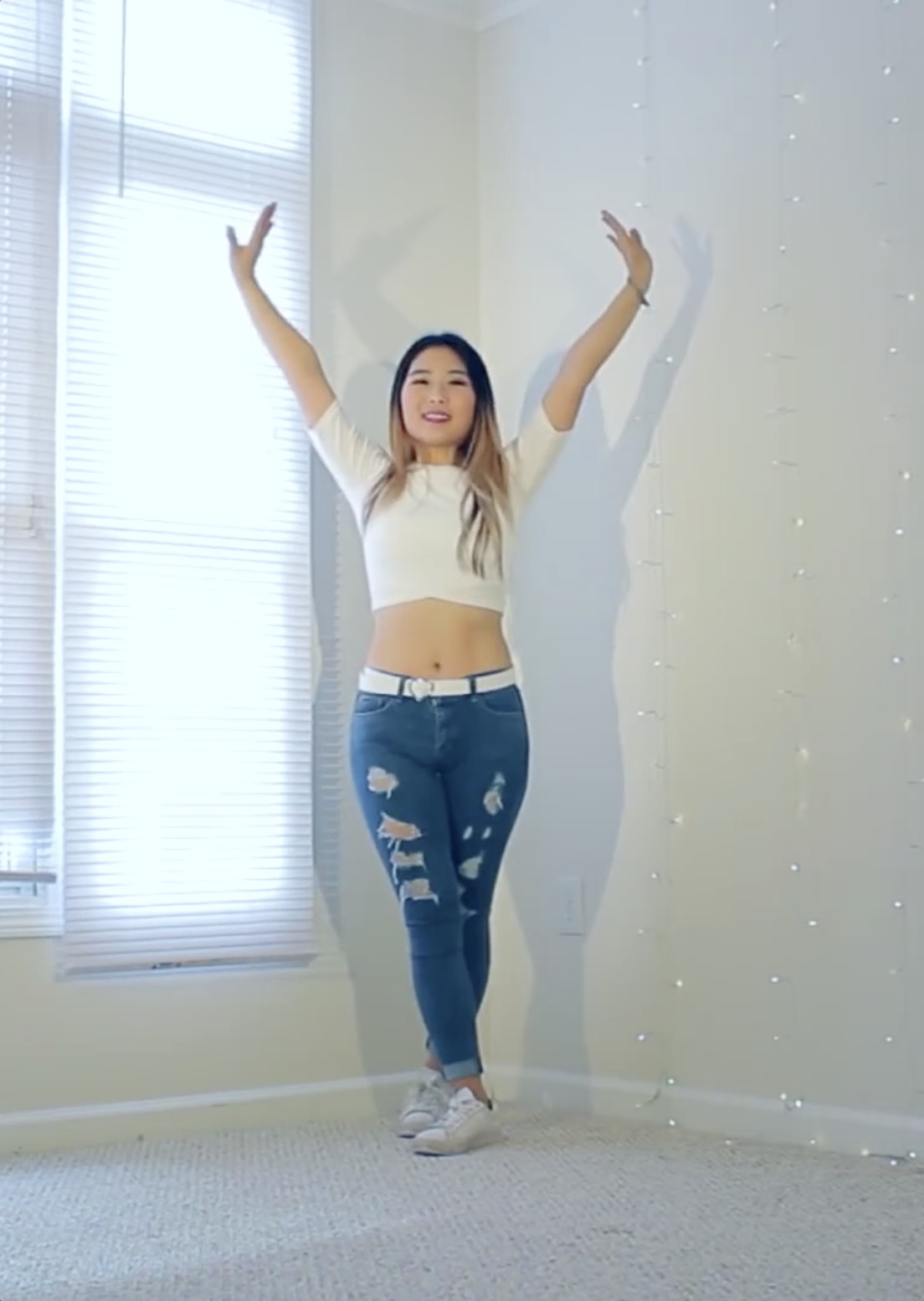}\\ 
    (a)&(b)&(c)&(d)&(e)\\
    \end{tabular}
    }
\resizebox{.95\textwidth}{!}{
    \begin{tabular}{cccccc}
    \includegraphics[width=.14835821345234\linewidth]{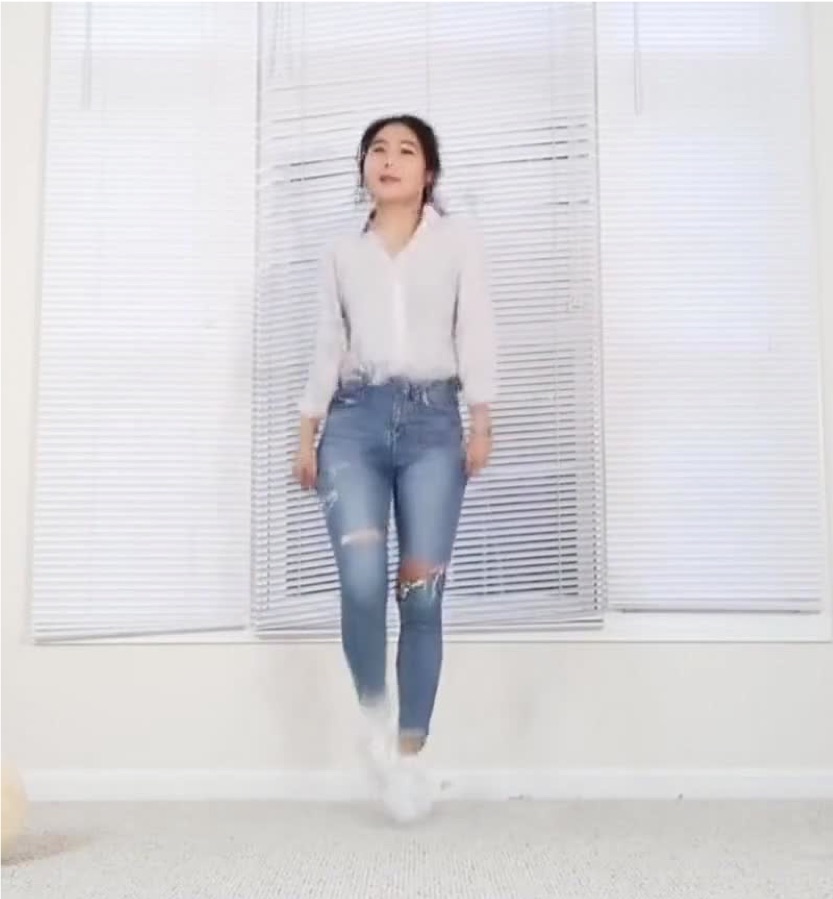}&
    \includegraphics[width=.14835821345234\linewidth]{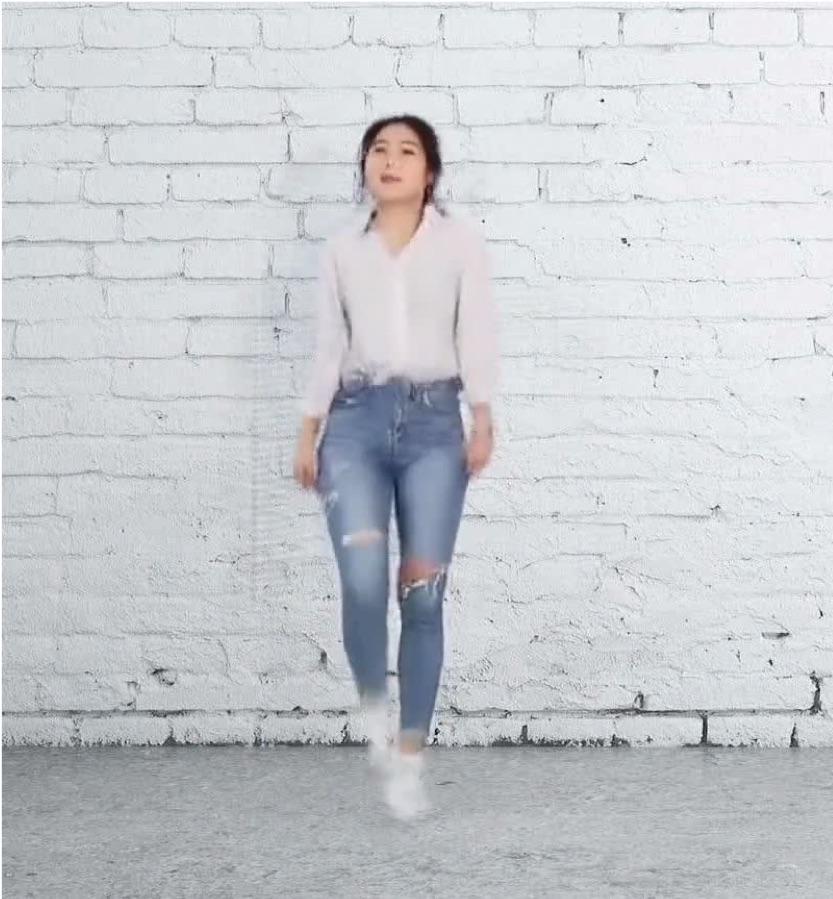}& 
    \includegraphics[width=.14835821345234\linewidth]{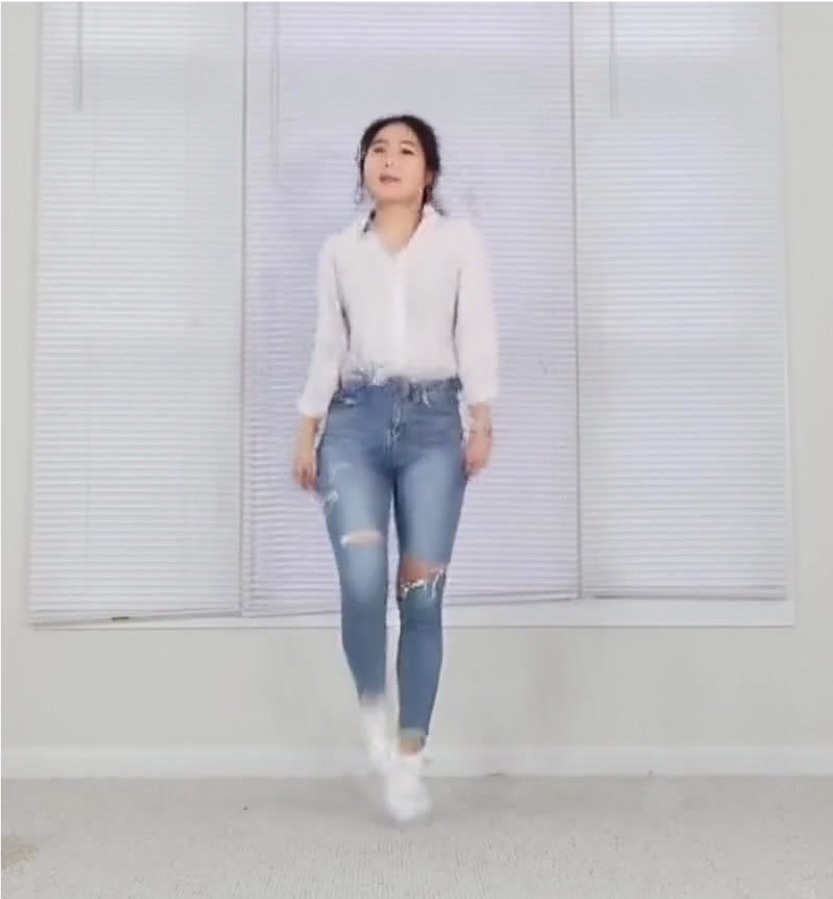}& 
    \includegraphics[width=.14835821345234\linewidth]{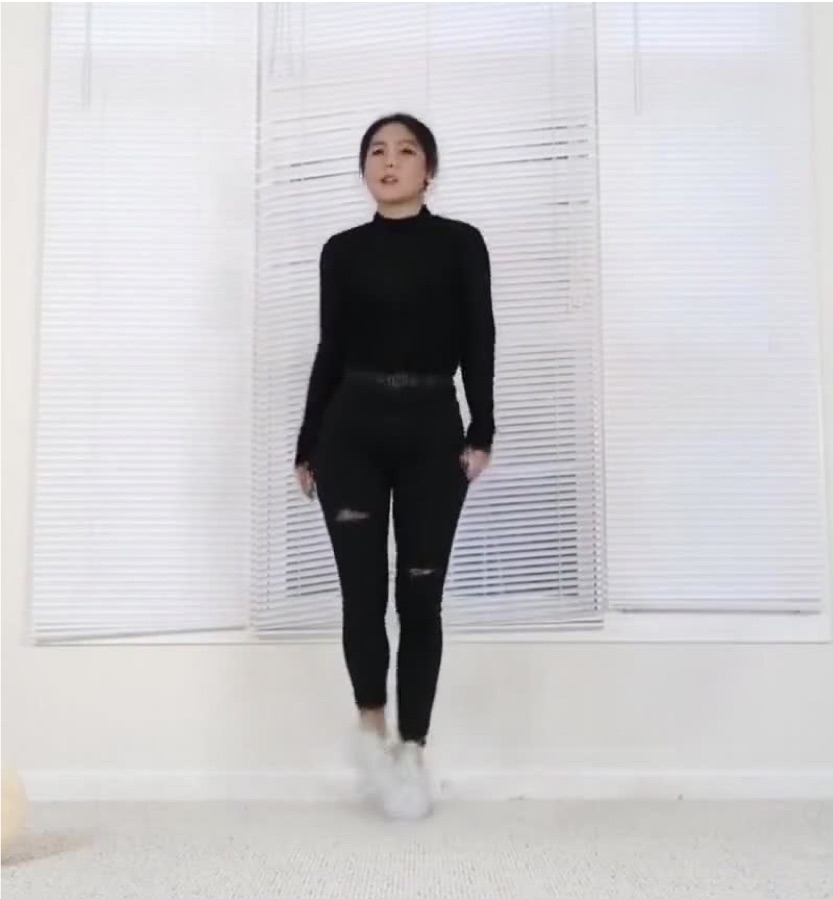}&
    \includegraphics[width=.14835821345234\linewidth]{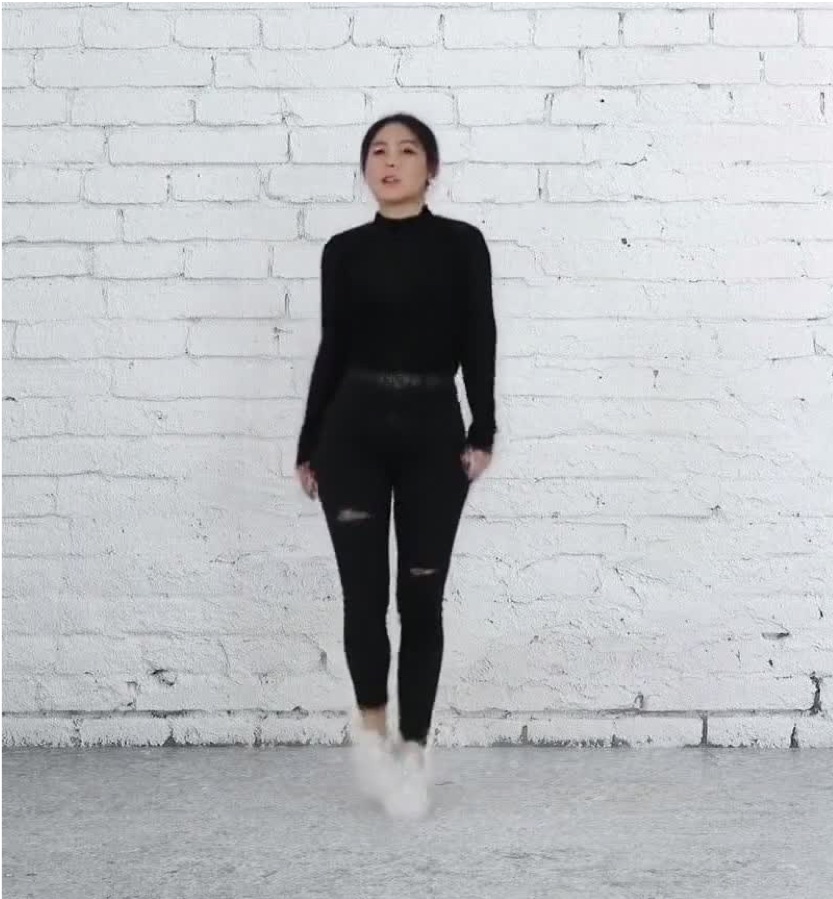}&
    \includegraphics[width=.14835821345234\linewidth]{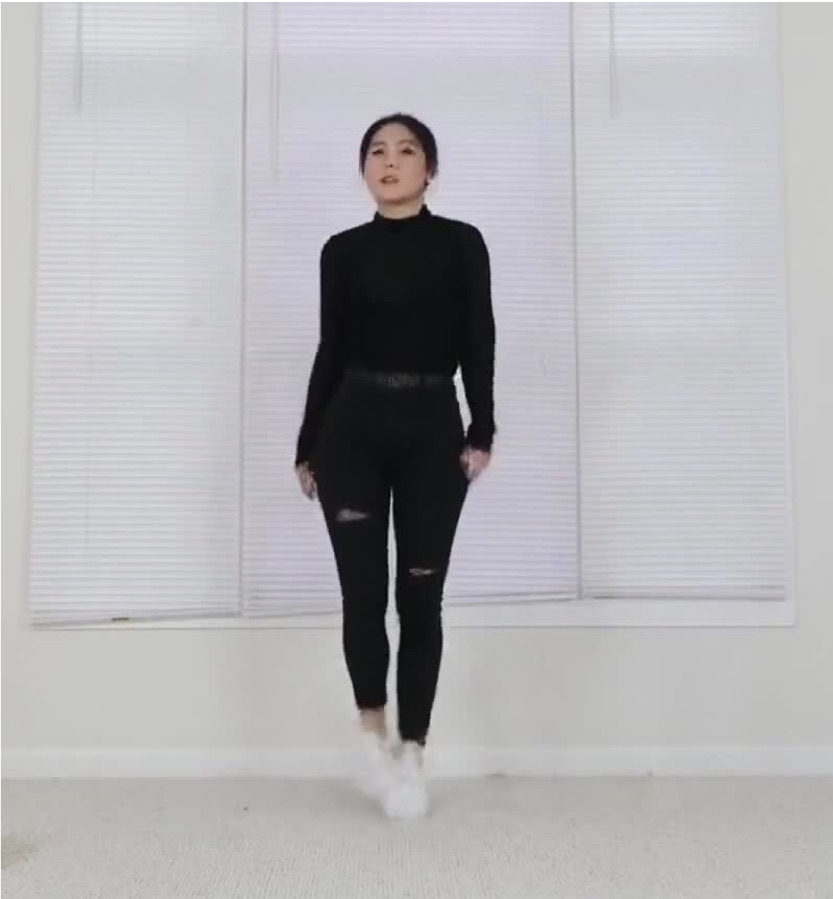}\\
    (f)&(g)&(h)&(i)&(j)&(k)\\
    \end{tabular}
}
        \caption{A comparison of the P2F network with the vid2vid method of~\cite{wang2018vid2vid}. (a) The target-pose image, (b) the pose extracted from this image, (c) the result of vid2vid, (d) our result, (e) a frame from the reference video. Many artifacts are apparent in the background produced by vid2vid. vid2vid also distorts the character's appearance and dimensions to better match the pose. (f-k) The same pose, displayed by two characters on three different backgrounds, demonstrates our advantage over vid2vid in replacing backgrounds.}
    \label{fig:vid2vid}
\end{figure*}

\section{Experiments}

The method was tested on multiple video sequences, see supplementary video\footnote{\url{https://youtu.be/sNp6HskavBE}}. The first video shows a tennis player outdoors, the second video, a person swiping a sword indoors, and the third, a person walking. The part of the videos used for training consists of 5.5min, 3.5min, and 7.5min, respectively. In addition, for comparative purposes, we trained the P2F network on a three min video of a dancer, which was part of the evaluation done by~\cite{wang2018vid2vid}.

The controllable output of the tennis player is shown in Fig.~\ref{fig:teaser}, which depicts the controller signal used to drive the pose, as well as the generated pose $p_i$, {\color{black} object $obj_i$}, mask $m_i$, raw frame $z_i$, and output frame $f_i$. A realistic character is generated with some artifacts (see supplementary video) around the tennis racket, for which the segmentation of the training video is only partially successful. Fig.~\ref{fig:backgrounds_tennis} depicts additional results, in which the character is placed on a diverse set of backgrounds containing considerable motion.
Fig.~\ref{fig:walk_v2} depicts a controlled walking character along with the control signal and the generated poses. A second walking sequence and a fencing sequence can be found in the appendix Fig.~\ref{fig:walk_v1}.

A comparison of the P2F network with the pix2pixHD method of~\cite{wang2018pix2pixHD} is provided in Tab.~\ref{tab:pix2pixhd_comp}, {\color{black}and as a figure in Fig.~\ref{fig:pix2pixhd}}. We compare by Structural Similarity (SSIM~\cite{wang2004image}) and Learned Perceptual Image Patch Similarity (LPIPS~\cite{zhang2018perceptual}) distance methods. The mean and standard deviation are calculated for each generated video. The LPIPS method provides a perceptual distance metric, by comparing the activations of three different network architectures (VGG~\cite{simonyan2014very}, AlexNet~\cite{krizhevsky2014one}, and SqueezeNet~\cite{iandola2016squeezenet}), with an additional linear layer set on top of each network. For each dataset, we select a test set that was not used during training. Although this test set is evaluated as the ground-truth, there is a domain shift between the training and the test video: the tennis test set contains dynamic elements, such as other players, crowd, and a slight difference in camera angle; the walking test set contains different character clothing, background lighting, and camera angle. The fencing test set is more similar to the training set.  As seen in the appendix, the baseline method results in many background and character artifacts, and a degradation in image and character quality, as it is forced to model the entire scene, rather than focus solely on the character and its shadow, as our method does. This is also apparent in the statistics reported in the table.

Another experiment dedicated to the P2F network (other methods do not employ P2P), compares it with the recent vid2vid method by~\cite{wang2018vid2vid}. The results are reported in the supplementary video and in Fig.~\ref{fig:vid2vid}(a-e). Shown are the target image from which the driving pose is extracted, the extracted pose, the results of the baseline method, and our result. As can be seen, our method handles the background in a way that creates far fewer distortions. The characters themselves are mostly comparable in quality, despite our choice not to add a dedicated treatment to the face. In addition, despite not using the temporal smoothness term, our method produces videos that are as smooth. Finally, the proportions of the character in our video are better maintained, while in the baseline model, the character is slightly distorted toward the driving pose. In addition, as we demonstrate in Fig.~\ref{fig:vid2vid}(f-k), our method has the ability to replace the background.

\noindent\textbf{Ablation study} We test the effect of several novel P2P network components, both by structural similarity (SSIM~\cite{wang2004image}) and perceptual patch similarity (LPIPS~\cite{zhang2018perceptual}) distance methods. The test is performed when predicting one frame into the future (in a ``teacher forcing'' mode). The results in Tab.~\ref{tab:ablation} demonstrate that our conditioning block is preferable to the conventional one, and that adding the object channel is beneficial. Selecting the model based on the minimal discriminator feature-matching loss is also helpful.

\section{Conclusion}

In this work, we develop a novel method for extracting a character from an uncontrolled video sequence and then reanimating it, on any background, according to a 2D control signal. Our method is able to create long sequences of coarsely-controlled poses in an autoregressive manner. These poses are then converted into a video sequence by a second network, in a way that enables the careful handling and replacement of the background, which is crucial for many applications. 
Our work paves the way for new types of realistic and personalized games, which can be casually created from everyday videos. In addition, controllable characters extracted from YouTube-like videos can find their place in the virtual worlds and augmented realities. 

\section*{Acknowledgments}
The authors would like to thank Lisa Rhee, Ilkka Hartikainen, and Adrian Bryant for allowing us to use their videos for training.

\bibliographystyle{ieee}
\bibliography{gans}

\clearpage
\appendix
\section{Additional Pose2Pose Network Architecture and Implementation Details}

We follow the naming convention of ~\cite{wang2018pix2pixHD,CycleGAN2017,perceptual}. Let Ck denote a Conv-InstanceNorm-ReLU
layer with k filters, each with a kernel size of 7x7, with a stride of 1. Dk denotes a Convolution-InstanceNorm-ReLU
layer with $k$ filters and a stride of 2, where reflection padding is used. Vk denotes a vanilla residual block with two 3x3 convolutional layers with the same number of filters on both layers. Wk denotes a conditioned residual block. Uk denotes a 3x3 Fractional-Strided-Convolution-InstanceNorm layer with $k$ filters, and a stride of $0.5$. 

The generator, i.e., the P2P network, can then be described as: C64, D128, D256, D512, D1024, V1024, W1024, 
W1024, W1024, W1024, W1024, W1024, W1024, V1024, 
U512, U256, U128, U64, C3.

The input images are scaled to a width size of 512 pixels, with the height scaled accordingly.

The discriminators are two PatchGANs ~\cite{pix2pix} with an identical architecture of C64,C128,C256,C512, working at the input resolution and a lower resolution, down-sampled by an average-2D-pooling operation with a kernel size of 3, and a stride of 2. 

The architecture of the P2F network is similar to that of the P2P network, with the following adjustments: (i) the conditional residual blocks are replaced by non residual ones, (ii) the input of P2F has 6 channels for $p_i$ and $p_{i-1}$, (iii) there is an additional head generating the mask output, which uses a sigmoid activation function.

\section{Additional images}

\begin{figure}
  \centering
  \begin{tabular}{ccc} 
  \includegraphics[width=2.5cm]{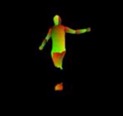} & \includegraphics[width=2.5cm]{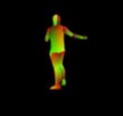} & \includegraphics[width=2.5cm]{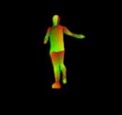} \\ 
  \includegraphics[width=2.5cm]{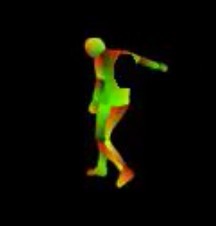} & \includegraphics[width=2.5cm]{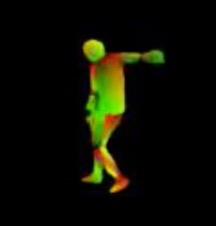} & \includegraphics[width=2.5cm]{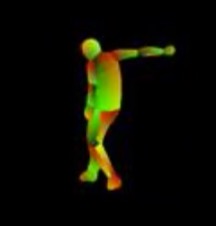} \\ 
  (a) & (b) & (c)
  \end{tabular}
  \caption{The occlusion-based augmentation technique used to increase robustness during training the P2P network. Each row is a single sample. (a) $p_{i-1}$ with part of it occluded by a random ellipse, (b) the predicted pose $\hat p_i$, (c) the ground truth pose $p_i$. The generated output seems to "fill in" the missing limbs, as well as predict the next frame.  In this figure and elsewhere, the colors represent the 3D UV mapping.} \label{fig:training_techniques}
\end{figure}

Fig. ~\ref{fig:training_techniques} depicts the random occlusion process (P2P training), in which a black ellipse of random size and location is added to each input pose frame within the detection bounding box. This results in an impaired pose, with characteristics that are similar to "naturally" occurring imperfections. 

\begin{figure*}
  \includegraphics[width=0.9148\textwidth]{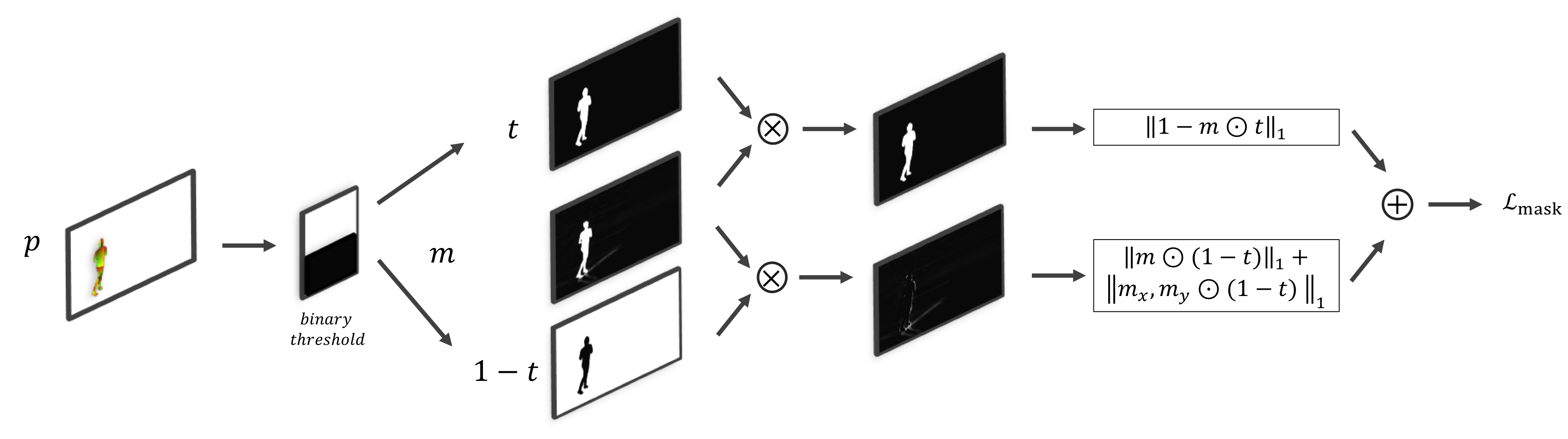}
  \caption{Mask losses applied during the P2F network training. An inverse binary-thresholded mask is used to penalize pixel intensity for both the generated mask and the raw output, in the regions excluding the character of interest. For the generated mask, we apply regularization over the derivatives in the x and y axes as well, to encourage smooth mask generation, and discourage high-frequency pattern generation.}
  \label{fig:regs}
\end{figure*}

The mask loss term $\mathcal{L}_{mask}$ of P2F (Sec.~5) is illustrated in Fig.~\ref{fig:regs}.

\begin{figure*}
  \includegraphics[width=0.848\textwidth]{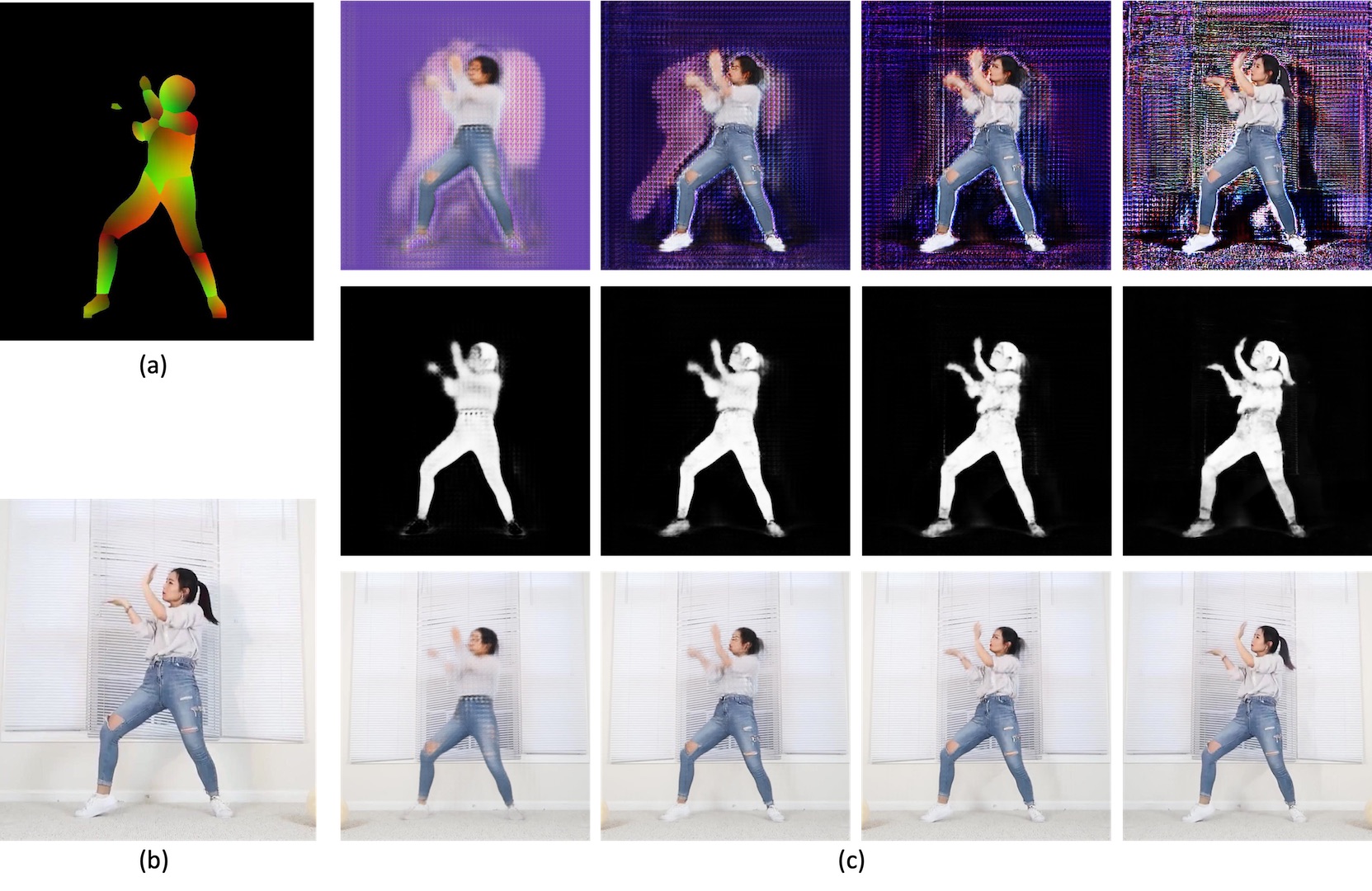}
  \caption{Training the P2F network. (a) a sample pose, (b) the target frame, (c) the generated raw frame, the mask, and the output frame at different epochs: 10, 30, 50, and 200 (final).}
  \label{fig:training_seq}
\end{figure*}

Fig.~\ref{fig:training_seq} depicts the progression during training of the pose-to-frame dancer model. As training progresses, the details of the dancer become sharper and the hair becomes part of the mask, despite being outside the DensePose detected area (i.e., off pixels in $t$).

\begin{figure*}
  \includegraphics[width=0.995\textwidth]{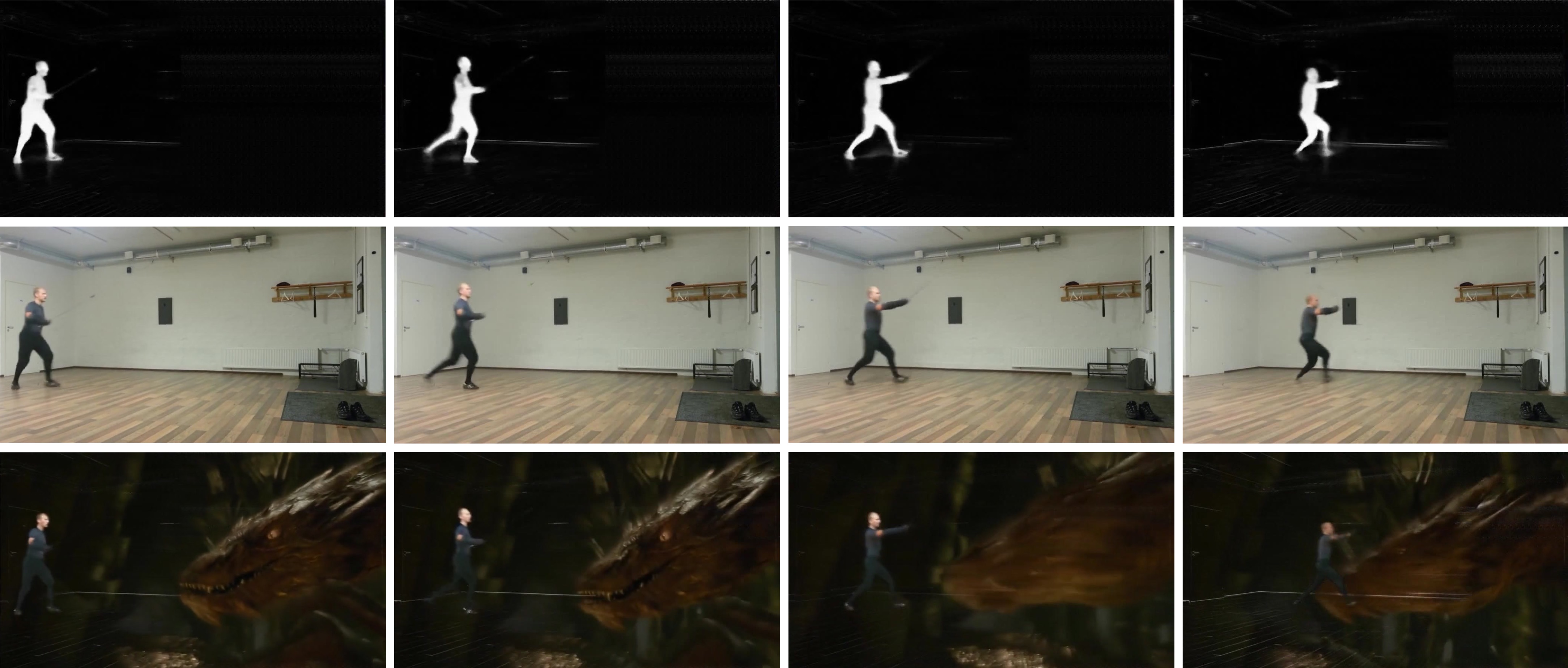}
  \caption{Generated frames for the controllable fencing character. Each column is a different pose. The rows are the obtained mask, and the placement on two different backgrounds: the one obtained by applying a median filter to the reference video and one taken from a motion picture.}
  \label{fig:fencing_bkgnds_v2}
\end{figure*}

\begin{figure*}
  \includegraphics[width=0.995\textwidth]{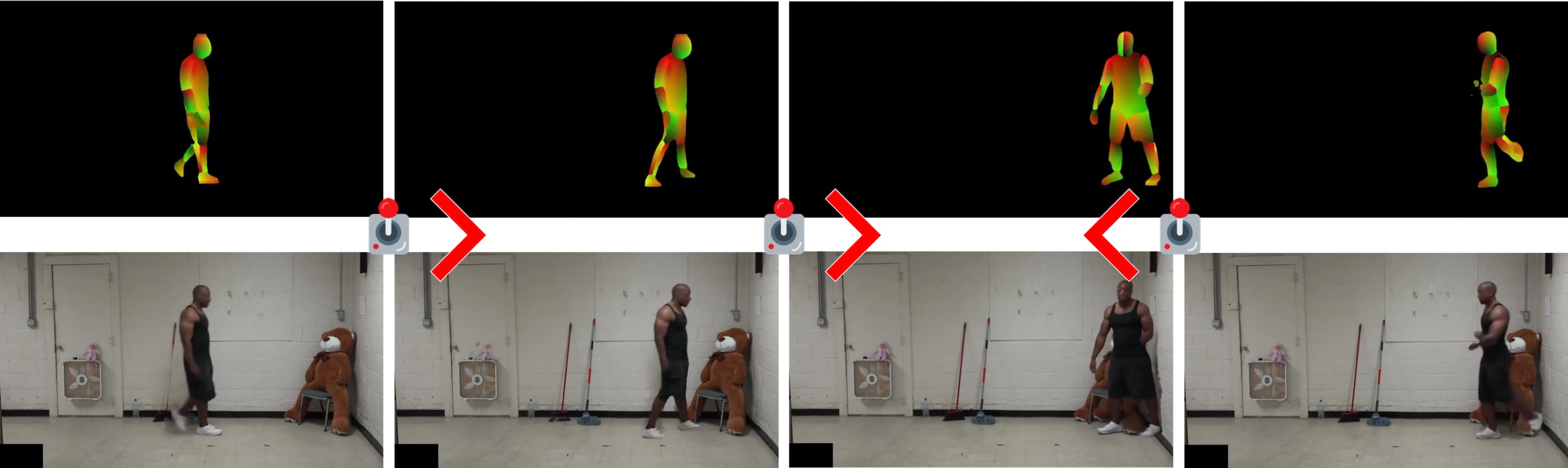}
  \caption{Synthesizing a walking character, emphasizing the control between the frames. Shown are the sequence of poses generated in an autoregressive manner, as well as the generated frames.}
  \label{fig:walk_v1}
\end{figure*}

The fencing character is shown in Fig.~\ref{fig:fencing_bkgnds_v2}. The mask for various frames in the controlled sequence is shown, as well as two backgrounds: the background of the reference video, and an animated background. Fig.~\ref{fig:walk_v1} depicts the controlled walking character along with the control signal and the generated poses.

\begin{figure*}
    \centering
    \begin{tabular}{ccc}
    \includegraphics[width=0.2967164269\linewidth]{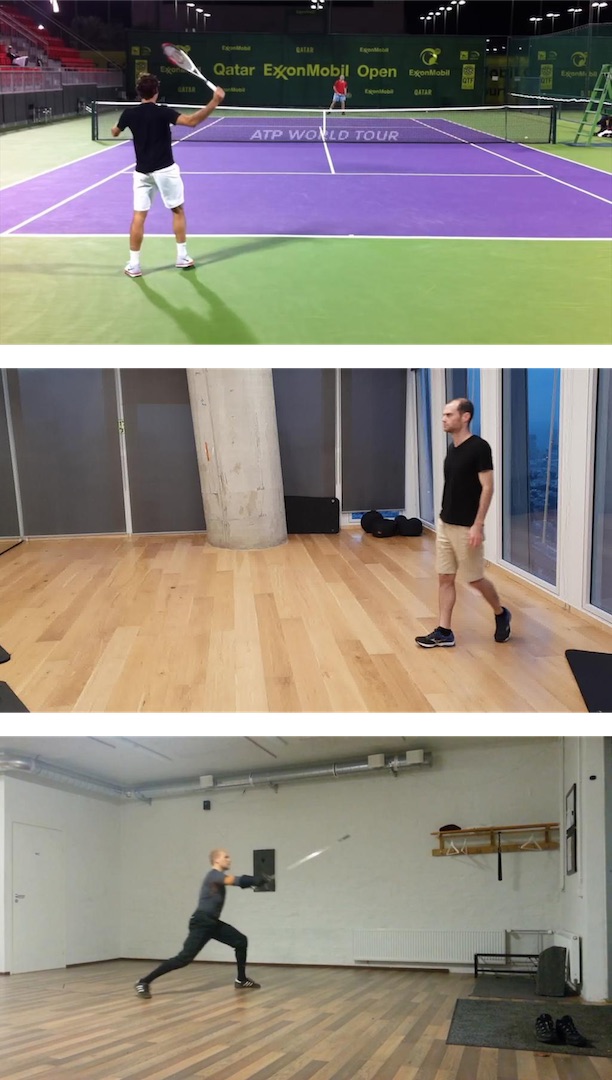}&
    \includegraphics[width=0.2967164269\linewidth]{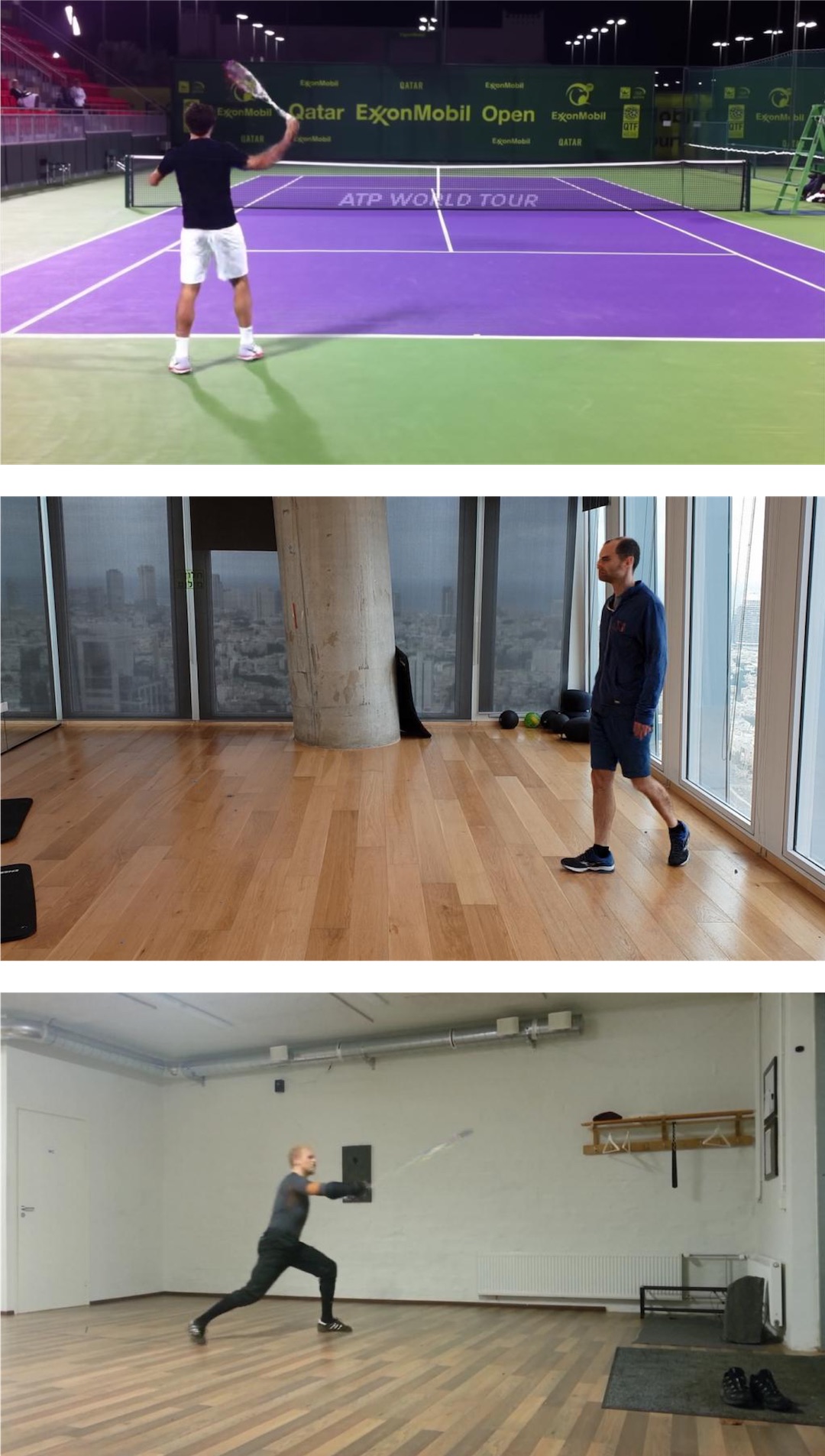}& 
    \includegraphics[width=0.2967164269\linewidth]{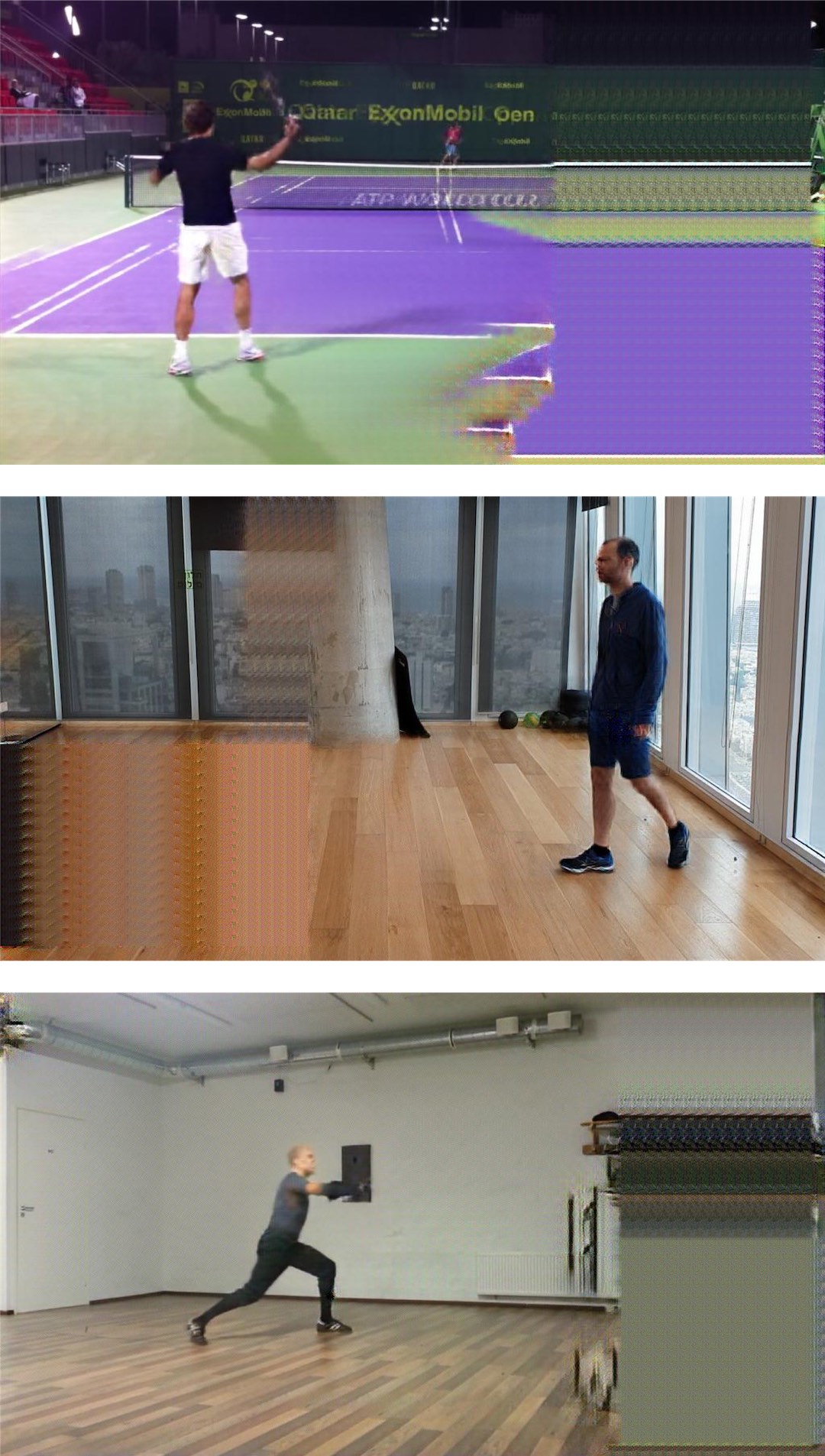}\\
    (a)&(b)&(c)\\
    \end{tabular}

        \caption{{\color{black}A comparison of the P2F network with the pix2pixHD method of~\cite{wang2018pix2pixHD}. (a) Ground truth image used as the pose source, (b) our result, (c) The results of pix2pixHD. The baseline method results in many background artifacts, as it generates the entire frame. The degradation in image quality is apparent as well, and that of the character in particular.}}
    \label{fig:pix2pixhd}
\end{figure*}

Fig~\ref{fig:pix2pixhd} compares visually with the baseline method of pix2pixHD~\cite{wang2018pix2pixHD}. As can be seen, the baseline method results in many background and character artifacts, a degradation in image and character quality, as it is forced to model the entire scene, rather than focusing solely on the character and the environmental factors, such as in our method. 

\end{document}